\documentclass[11pt]{article}

\usepackage[preprint]{acl}
\usepackage{diagbox}
\usepackage{makecell}
\usepackage{placeins}
\usepackage{float}

\usepackage{times}
\usepackage{latexsym}

\usepackage[T1]{fontenc}

\usepackage[utf8]{inputenc}

\usepackage{microtype}

\usepackage{inconsolata}

\usepackage{graphicx}

\usepackage{xcolor}
\usepackage{tcolorbox}
\usepackage{adjustbox}
\usepackage{amssymb}
\usepackage{booktabs}
\usepackage[table]{xcolor}
\usepackage{amsmath}

\usepackage{tikz}
\usetikzlibrary{arrows.meta,positioning,fit,calc,shadows,backgrounds,shapes.geometric}
\usepackage{pgfplots}
\usepgfplotslibrary{groupplots,colormaps}
\pgfplotsset{compat=1.18}
\pgfdeclarelayer{background}
\pgfsetlayers{background,main}
\usepackage{colortbl}
\usepackage{subcaption}
\usepackage{venndiagram}
\usepackage{amssymb}
\usepackage{amsmath}
\usepackage{booktabs}
\usepackage{multirow}
\usepackage{array}
\usepackage{sansmath} 

\pgfplotsset{
  colormap={gainmap}{
    rgb255=( 49, 54,149)   
    rgb255=( 69,117,180)
    rgb255=(116,173,209)
    rgb255=(171,217,233)
    rgb255=(224,243,248)
    rgb255=(255,255,255)  
    rgb255=(254,224,144)
    rgb255=(253,174, 97)
    rgb255=(244,109, 67)
    rgb255=(215, 48, 39)
    rgb255=(165,  0, 38)  
  }
}

\pgfplotsset{
    colormap={softgray}{
    rgb255(0cm)=(255,255,255)
    rgb255(1cm)=(240,240,240)
    rgb255(2cm)=(200,200,200)
    rgb255(3cm)=(150,150,150)
    rgb255(4cm)=(100,100,100)
}
}
\title{Parametric Knowledge in RAG-SFT\\ for Domain-Specific Document Generation} 

\author{
\textbf{Julian Oestreich\textsuperscript{1}\thanks{Equal contribution.}},
\textbf{Maximilian Bley\textsuperscript{1}\footnotemark[1]},
\textbf{Frank Binder\textsuperscript{1}},
\textbf{Lydia Müller\textsuperscript{1}}
\\
\textbf{André Alcalde\textsuperscript{2}},
\textbf{Maksym Sydorenko\textsuperscript{2}}
\\
\\
 \textsuperscript{1}Institute for Applied Informatics (InfAI) at Leipzig University
 \\
 \textsuperscript{2}CELUS GmbH, Munich
\\
 \texttt{\{oestreich, maximilian.bley, binder, lydia.mueller\}\href{mailto:email@domain}{@infai.org}}
 \\
 \texttt{\{andre.alcalde, max.sydorenko\}\href{mailto:email@domain}{@celus.io}}
}

\begin{document}
\maketitle

\begin{abstract}

Retrieval-Augmented Generation (RAG) fine-tuning has shown substantial improvements over vanilla RAG, yet most studies target document question answering, leaving open whether these gains transfer to specialized tasks. 
We study supervised RAG fine-tuning (RAG-SFT) for requirements document generation in the electronics engineering domain, adapting two 7B models under two different training data strategies.
Because Rouge and BertScore poorly capture factuality on long technical text, we introduce \textsc{C-FEX}, a claim-based evaluation pipeline that attributes each response claim to its origin (augmented prompt or reference response), and propose Parametric Knowledge Precision (PKP), which isolates claims originating from the model's weights and measures their correctness. 
We show that a prior metric to assess parametric knowledge decomposes as PKP $\times$ PR, separating the rate of parametric output (PR) from its quality (PKP).
Empirically, fine-tuned 7B models match or exceed a 72B baseline; standard metrics disagree with claim-based factuality and can mislead about fine-tuning gains; and, fine-tuning does not reinforce correct parametric knowledge but suppresses hallucination---models speak from their weights less often but far more reliably. 
%
%
\end{abstract}
\begin{figure*}[t]
    \centering
    \resizebox{\linewidth}{!}{
\definecolor{srcfill}{HTML}{E8F0FE}
\definecolor{srcstroke}{HTML}{6C8EBF}
\definecolor{dbfill}{HTML}{FFF2CC}
\definecolor{dbstroke}{HTML}{D6B656}
\definecolor{llmfill}{HTML}{D5E8D4}
\definecolor{llmstroke}{HTML}{4F8A4C}
\definecolor{outfill}{HTML}{E1D5E7}
\definecolor{outstroke}{HTML}{9673A6}
\definecolor{panelfill}{HTML}{FAFAFA}
\definecolor{panelstroke}{HTML}{BFBFBF}
\definecolor{codegray}{HTML}{555555}
\definecolor{vR}{HTML}{4F8A4C}  
\definecolor{vC}{HTML}{C0392B}  
\definecolor{vU}{HTML}{2E6DA4}  
\begin{tikzpicture}[
  font=\sffamily,
  >={Latex[length=2.6mm]},
  node distance=6mm,
  panel/.style={
    rounded corners=5mm, draw=panelstroke, line width=1pt,
    fill=panelfill, inner sep=14pt, minimum height=13.2cm
  },
  paneltitle/.style={font=\sffamily\bfseries\Large, anchor=north},
  srcbox/.style={
    rounded corners=2.5mm, draw=srcstroke, line width=1pt,
    fill=srcfill, inner sep=8pt, align=left
  },
  outbox/.style={
    rounded corners=2.5mm, draw=outstroke, line width=1pt,
    fill=outfill, inner sep=8pt, align=left
  },
  llmnode/.style={
    rounded corners=2mm, draw=llmstroke, line width=1.1pt,
    fill=llmfill, inner sep=8pt, font=\sffamily\bfseries,
    align=center, minimum height=9mm
  },
  dbcyl/.style={
    cylinder, shape border rotate=90, aspect=0.28,
    draw=dbstroke, line width=1pt, fill=dbfill,
    minimum width=1.05cm, minimum height=1.15cm, inner sep=1pt,
    font=\sffamily\bfseries\small
  },
  arrow/.style={->, line width=1.3pt, draw=black!65},
  thinarrow/.style={->, line width=1pt, draw=black!55},
  dataarrow/.style={->, line width=1.8pt, draw=srcstroke, dashed},
  lbl/.style={font=\sffamily\bfseries\footnotesize},
  capt/.style={font=\sffamily\footnotesize\itshape, text=black!60},
]

\node[panel, minimum width=7.6cm] (panelA) at (0,0) {};
\node[paneltitle] at ($(panelA.north)+(0,-0.25cm)$) {\textbf{(A)} Claim Extraction};

\node[srcbox, minimum width=6.4cm, anchor=north, text width=5.9cm]
  (input) at ($(panelA.north)+(0,-1.25cm)$)
{%
\centerline{\bfseries\footnotesize Generated Response $j$}\\[+0.9mm]
{\scriptsize\ttfamily\color{codegray}
"text": "**Power Conversion Module**: This module is responsible for converting the input DC voltage (ranging from 300 to 400 volts) into a 3-phase AC output. It is designed to handle an output power range of 30 to 150 watts."\\
\centerline{\ldots}
}
};
\node[llmnode, below=6.5mm of input] (extractLLM)
  {LLM-Extractor}; 
\draw[arrow] (input) -- (extractLLM);

\node[outbox, below=6.5mm of extractLLM, minimum width=6.4cm, text width=5.9cm]
  (claims)
{%
\centerline{ \bfseries\footnotesize Extracted Claims $\hat{\mathcal{C}}_{\mathcal{G}}$ $=\{c_{1j}, ... c_{nj}\}$}\\[+1.1mm]
{\scriptsize\ttfamily\color{codegray}
$c_{1j}$: "The power conversion module of the PowerWave Inverter Platform converts DC input to a 3-phase AC output."
\\[0.6mm]
$c_{2j}$: "The power conversion module of the PowerWave Inverter Platform supports input voltages ranging from 300 to 400 V DC."
\\[0.6mm]
$c_{3j}$: "The power conversion module of the PowerWave Inverter Platform has an adjustable output power between 30 to 150 W."
\\[0.6mm]
$c_{4j}$: "The PowerWave Inverter Platform is intended for powering home appliances including refrigerators, air conditioners, and washing machines."
\\[0.6mm]
$c_{5j}$: "The protection and safety module of the PowerWave Inverter Platform incorporates short-circuit protection."
\\[0.6mm]
\centerline{\ldots}
}%
};
\draw[arrow] (extractLLM) -- (claims);

\node[panel, minimum width=7.6cm, right=1.0cm of panelA] (panelB) {};
\node[paneltitle] at ($(panelB.north)+(0,-0.25cm)$) {\textbf{(B)} Attribution};

\node[outbox, anchor=north, minimum width=6.4cm, text width=5.9cm, align=left]
  (query) at ($(panelB.north)+(0,-1.25cm)$)
{%
\centerline{\bfseries\footnotesize Candidate Query $c_{ij} \in$ $\hat{\mathcal{C}}_{\mathcal{G}}$}\\[0.9mm]
{\scriptsize\ttfamily\color{codegray}
$c_{1j}$: "The power conversion module of the PowerWave Inverter Platform converts DC input to a 3-phase AC output."\\[0.6mm]
}%
};

\node[dbcyl, below=8mm of query, xshift=-2.15cm] (dbR) { $\mathcal{C}_{\mathcal{R}}$ };
\node[dbcyl, below=8mm of query]                 (dbC) { $\mathcal{C}_{\mathcal{C}}$ };
\node[dbcyl, below=8mm of query, xshift= 2.15cm] (dbU) { $\mathcal{C}_{\mathcal{U}}$ };
\draw[arrow] (query) -- (dbC);

\node[srcbox, below=7mm of dbC, minimum width=6.4cm, text width=5.9cm, align=left]
  (cands)
{%
\centerline{\bfseries\footnotesize Top-$k$ Similarity Candidates}\\[+1.1mm]
{\scriptsize\ttfamily\color{codegray}
reference@1 (sim $=0.8550$):\\"The Power Conversion Module converts input DC voltage (300 to 400 volts) into a 3-phase AC output."\\[0.6mm]
context@1 (sim $=0.5562$):\\"RD-FSB50450A performs DC to DC conversion."\\[0.6mm]
user\_input@1 (sim $=0.9369$):\\"The PowerWave Inverter Platform converts DC power into a stable 3-phase AC output."\\[0.6mm]
}%
};
\draw[arrow] (dbR.south) -- (cands.north);
\draw[arrow] (dbC.south) -- (cands.north);
\draw[arrow] (dbU.south) -- (cands.north);

\node[llmnode, below=6mm of cands, text width=4.6cm] (judge)
  {LLM-as-a-Judge};
\draw[arrow] (cands) -- (judge);

\node[outbox, below=6mm of judge, minimum width=6.4cm, text width=5.9cm, align=left]
  (attr)
{%
\centerline{\bfseries\footnotesize Attribution $A_{ij}$}\\[0.9mm]
{\scriptsize\ttfamily\color{codegray}
"sources": ["reference", "user\_input"]\\[0.6mm]
}
};
\draw[arrow] (judge) -- (attr);

\node[panel, minimum width=7.0cm, right=1.0cm of panelB] (panelC) {};
\node[paneltitle] at ($(panelC.north)+(0,-0.25cm)$) {\textbf{(C)} Metrics};

\node[outbox, anchor=north, minimum width=5.6cm, text width=5.1cm, align=center]
  (allAttr) at ($(panelC.north)+(0,-1.25cm)$)
{%
{\bfseries\footnotesize All Attributed Claims}\\[0.6mm]
{\scriptsize $\{(c_{1j},A_{1j}),(c_{2j},A_{2j}),\ldots,(c_{nj},A_{nj})\}$}%
};

\begin{scope}[shift={($(allAttr.south)+(0,-3.05cm)$)}]
  \def\rad{1.45cm}
  \node[draw=black!45, line width=1pt, dashed, rounded corners=3mm,
        minimum width=6.0cm, minimum height=5.2cm] (Gset) at (0,-0.15) {};
  \node[font=\sffamily\bfseries\footnotesize, anchor=north west]
        at ($(Gset.north west)+(0.12,-0.08)$) {$\hat{\mathcal{C}}_{\mathcal{G}}$};
  \begin{scope}[opacity=0.45, blend mode=multiply]
    \fill[vR] (90:0.95cm) circle (\rad);
    \fill[vU] (210:0.95cm) circle (\rad);
    \fill[vC] (330:0.95cm) circle (\rad);
  \end{scope}
  \draw[vR, line width=1pt] (90:0.95cm) circle (\rad);
  \draw[vU, line width=1pt] (210:0.95cm) circle (\rad);
  \draw[vC, line width=1pt] (330:0.95cm) circle (\rad);
  \node[vR, font=\sffamily\bfseries] at (90:1.5cm) { $\mathcal{S}_{\mathcal{R}}$ };
  \node[vU, font=\sffamily\bfseries] at (230:1.75cm) { $\mathcal{S}_{\mathcal{U}}$ };
  \node[vC, font=\sffamily\bfseries] at (350:1.75cm) {  $\mathcal{S}_{\mathcal{C}}$ };
  \node[font=\sffamily\footnotesize\itshape, text=black!60]
        at (0,-2.25) {$\mathcal{H}= \hat{\mathcal{C}}_{\mathcal{G}}\setminus( \mathcal{S}_{\mathcal{R}} \cup \mathcal{S}_{\mathcal{C}} \cup \mathcal{S}_{\mathcal{U}})$};
\end{scope}

\coordinate (vennbottom) at ($(allAttr.south)+(0,-6.35cm)$);

\node[srcbox, anchor=north, minimum width=5.9cm, text width=5.4cm, align=center]
  (formula) at (vennbottom)
{%
{\bfseries\footnotesize Parametric Knowledge Precision}\\[1.6mm]
$\displaystyle \mathrm{PKP}=\frac{|\mathcal{P}_{ref}|}{|\mathcal{P}_{ref}|+|\mathcal{H}|}$\\[2.6mm]
{\footnotesize
$\mathcal{P}_{ref} = \mathcal{S}_{\mathcal{R}}\setminus(\mathcal{S}_{\mathcal{C}}\cup \mathcal{S}_{\mathcal{U}})$\\[1.0mm]
$\mathcal{H} = \hat{\mathcal{C}}_{\mathcal{G}}\setminus( \mathcal{S}_{\mathcal{R}} \cup \mathcal{S}_{\mathcal{C}} \cup \mathcal{S}_{\mathcal{U}})$
}%
};

\draw[dataarrow] ([yshift=2mm]panelA.east) -- node[lbl, above, text=srcstroke]{$\hat{\mathcal{C}}_{\mathcal{G}}$} ([yshift=2mm]panelB.west);
\draw[dataarrow] ([yshift=2mm]panelB.east) -- node[lbl, above, text=srcstroke]{$(c_{ij},A_{ij})$} ([yshift=2mm]panelC.west);

\end{tikzpicture}
    }
    \caption{\textsc{C-FEX}: Claim Extraction + Fact Evaluation. \textbf{(A) Extraction:} For each example, claims are extracted from the generated response ($\hat{\mathcal{C}}_{\mathcal{G}}$) and its sources (reference $\mathcal{C}_{\mathcal{R}}$, context $\mathcal{C}_{\mathcal{C}}$, and user query $\mathcal{C}_{\mathcal{U}}$). \textbf{(B) Attribution:} Each generated claim $\hat{\mathcal{C}}_{\mathcal{G}}$ is attributed to a source by matching against top-k candidate claims retrieved from each source. \textbf{(C) Metrics:} Claims and their attributions are aggregated for metric computations. E.g. excluding response claims attributed to the user query or context ($\mathcal{S}_{\mathcal{U}}$ and $\mathcal{S}_{\mathcal{C}}$) allows computation of \textbf{Parametric Knowledge Precision (PKP)} to investigate the quality of parametric knowledge ($\mathcal{P_{\textit{ref}}}+\mathcal{H}$) in different RAG-Generators.}
    \label{fig:placeholder}
\end{figure*}
\section{Introduction}
Retrieval-Augmented Generation (RAG) has emerged as a popular training-free adaptation approach by enhancing language models with external knowledge\,\cite{gao2023ragsurvey}. 
Concurrently, fine-tuning in RAG settings has led to substantial improvements over vanilla RAG, including better task solving capabilities, and increased robustness to retrieval errors\,\cite{wang-etal-2024-searching,liu2024chatqa,yoranmaking2024,zhang2024raft,bhushan-etal-2025-parag,xuchatqa2,lin2025resisting}.
Although considerable research on RAG fine-tuning has been published, most of the studies focus on arguably the most common RAG application, question answering over document corpora, and it remains uncertain whether their findings transfer to specialized tasks too.
%
Furthermore, the rather short answers in document Q\&A are often evaluated using standard NLP metrics, such as Rouge or BertScore, which are well known to exhibit shortcomings in detecting hallucinations\,\cite{kaster-etal-2021-global,honovich-etal-2022-true-evaluating,jiang-etal-2025-towards-better}, particularly for longer text\,\cite{wei2024long,samarinas-etal-2025-beyond}. 
Recent metrics address these issues by transforming text into structured formats and leveraging LLMs as judges\,\cite{min-etal-2023-factscore,ru2024ragchecker,hu-etal-2024-knowledge,pradeep2025great,metropolitansky-larson-2025-towards}, but relatively little of this evaluation effort has focused on specialized generations in RAG settings. 
%

In this work, we study supervised RAG fine-tuning (RAF-SFT) for a task beyond question answering over document corpora: long-form document generation for requirements engineering in the electronics domain, where a model is prompted to expand an underspecified user query into a factually dense document by integrating retrieved context. 
%
Using a novel synthetic dataset, we adapt two 7B models under two data strategies---training on relevant context only, and contrastively on relevant and irrelevant context---and compare them against a 72B vanilla-RAG baseline under both successful retrieval (relevant chunks) and failed retrieval (only irrelevant chunks).
%
%
Since our RAG responses are long text generations and therefore hard to evaluate with standard NLP metrics, we transform them into a claim-based format, where each claim represents a single requirements statement.
Alongside this, we extract claims from each example's corresponding sources, namely the augmented prompt and the reference document. 
This allows us to trace back each response claim to its origin in order to evaluate different quality dimensions. 
We compute reference-backed metrics (e.g., precision and recall) as well as additional precision-side measurements, such as hallucination rate.
For evaluation, we use our own pipeline \textsc{C-FEX}, which integrates all steps into one framework (\autoref{fig:placeholder}). 
%
Building on \citet{ru2024ragchecker}'s Self-Knowledge (SK), we introduce Parametric Knowledge Precision (PKP), which removes claims attributable to the augmented prompt from its denominator before measuring correctness, thereby isolating knowledge that originates from the model's weights rather than confounding it with copied content (\textit{Metrics} in \autoref{fig:placeholder}). 
We further show that SK factorizes as SK = PKP $\times$ PR, separating how often a model answers from its weights (the parametric rate, PR) from how reliably it does so (PKP).
We investigate the following research questions:
\begin{description}
  \item[\textbf{ RQ1} ] Does RAG-SFT transfer to long-form requirements-document generation across different base models and retrieval conditions?
  \item[\textbf{ RQ2} ] How do standard NLP metrics compare to claim-based factuality when evaluating fine-tuning effects on long technical documents?
  \item[\textbf{ RQ3} ] How often do models go beyond the supplied information in the augmented prompt and recall information from their weights, how reliable is it, and how does RAG-SFT change this?
\end{description}
\paragraph{Contributions}
(1)~We propose \textsc{C-FEX}, a claim-based evaluation pipeline, comprising llm-based claim extraction, source attribution, and several metric computations, and release it publicly.\footnote{Code will be published upon acceptance.}
(2)~We introduce Parametric Knowledge Precision (PKP), which isolates the correctness of claims drawn from the model's weights, and show that \citet{ru2024ragchecker}'s Self-Knowledge factorizes as SK = PKP $\times$ PR, separating the rate of parametric output (PR) from its quality (PKP). 
(3)~We conduct an empirical study of RAG-SFT on a specialized task, showing that 7B adapters match or exceed a 72B baseline, that standard metrics can mislead about fine-tuning gains, and that fine-tuning suppresses hallucination rather than reinforcing correct parametric knowledge. 
%
%
\section{Related Work}
Adapting pretrained LLMs to new domains is well studied \citep{cheng2024adapting,tian2024finetuning,kujanp2025efficient}, though often framed as a choice between retrieval-augmented generation and fine-tuning \citep{ovadia-etal-2024-fine}.
Separately, substantial research has explored fine-tuning for RAG in open-domain settings, e.g., enhancing capabilities such as reading comprehension and long-context understanding\,\citep{liu2024chatqa, xuchatqa2}, or improving robustness through training on mixed relevant and irrelevant context\,\citep{yoranmaking2024}.
Most related to our work are approaches which unify these directions by training for domain-specific RAG.
For example, \citet{zhang2024raft} (RAFT) fine-tune on question-answer pairs, combined with noisy context (relevant and irrelevant chunks) and chain-of-thought answers, demonstrating great improvements over vanilla RAG across three domains: encyclopedic (Wikipedia), technical (Code Documentation) and medical Q\&A.
%
%
\citet{bhushan-etal-2025-parag} (PA-RAG) build on RAFT and \citet{yoranmaking2024}'s trainings recipe by generating novel, synthetic Q\&As maximizing document coverage and fine-tuning on those pairs augmented with relevant and irrelevant context, thereby improving both answer correctness and domain-groundedness.
%

Notably, the aforementioned research frames RAG almost exclusively as question answering over documents. This work addresses a different setting: a specialized domain with task-specific outputs beyond conventional Q\&A, where user queries must be integrated with retrieved context to produce comprehensive and lengthy responses. 

Related work on evaluating long-form text generation, motivated by the goal of better assessing factuality, structures outputs into fine-grained units such as sentences, subsentences or subject-predicate-object triples and compares them against a knowledge source, e.g., a reference text or Wikipedia\,\cite{thorne-etal-2018-fever,honovich-etal-2022-true-evaluating,hu-etal-2024-knowledge,jiang-etal-2025-towards-better,metropolitansky-larson-2025-towards}.
Among them, \citet{min-etal-2023-factscore} propose subsentences as the unit of evaluation and define correctness relative to a given knowledge source, rather then linking them to a external knowledge base. 
%
\citet{jiang-etal-2025-towards-better} propose a domain-specific LLM-based metric, manually verified by domain-experts, since standard NLP metrics (e.g., n-gram and embedding-based) are highly insufficient to capture the language of patent claim generations.
In RAG settings with long-form responses, \citet{pradeep2025great} create must-have response information given a query and context, then use an LLM to judge whether it appears in the response, similar to the RAGAS framework\,\cite{es-etal-2024-ragas}, which also uses LLMs for reference-free evaluation.

Since we have gold references, most related to our work is \citet{ru2024ragchecker}'s approach, which applies automatic triple-based evaluation comparing RAG answers to various sources. 
Using 70B LLMs, they extract triples from responses, references, and context chunks, then check entailment to compute precision, recall, and RAG-specific metrics based on triple provenance. 
Interestingly, they also study human correlations of the concepts of correctness\,/\,coverage of a response with their triple-based Precision and Recall metrics, and show, that they surpass not only Rouge and BertScore, but also popular other approaches like RAGAS (Table 5 in \citet{ru2024ragchecker}). 
They further introduce the Self-Knowledge metric, which we extend methodologically.
%
%
%
\section{Methodology}\label{sec:methods}

We study RAG-SFT for our use case by fine-tuning models on two different datasets and evaluating generated outputs against base models, using both standard NLP and claim-based metrics.
We define parametric knowledge in RAG, as the information in model answers, which is \textit{exclusively} originating from the model weights. 
To operationalize this, we segment text into full sentences, which should represent an atomic, decontextualized, verifiable, faithful and unambiguous statement, a \textit{claim} (inspired by \citet{metropolitansky-larson-2025-towards}).
The claims origin is judged, and claims are isolated which are not originating from the augmented prompt: We argue that in RAG settings claims in the prompt---e.g. from the retrieved context chunks---are more likely to be copied than recalled from the model weights.
Therefore we treat the remaining part of the response as a lower bound on \textit{parametric knowledge}.\footnote{Note that we borrow this definition from related work, which often measures domain-knowledge via Q\&A, therefore equate model knowledge with information in the answer.}
Since we use reference documents as the ground truth factual basis, we call claims, which are supported by the corresponding reference, \textit{parametric reference facts} ($\mathcal{P}_{ref}$), and unsupported claims \textit{hallucinations} ($\mathcal{H}$) (\autoref{fig:placeholder}).
%
\subsection{Data}
\begin{figure*}[t]
\centering
\begin{subfigure}[t]{0.48\textwidth}
    \centering
    \fbox{\includegraphics[width=\linewidth]{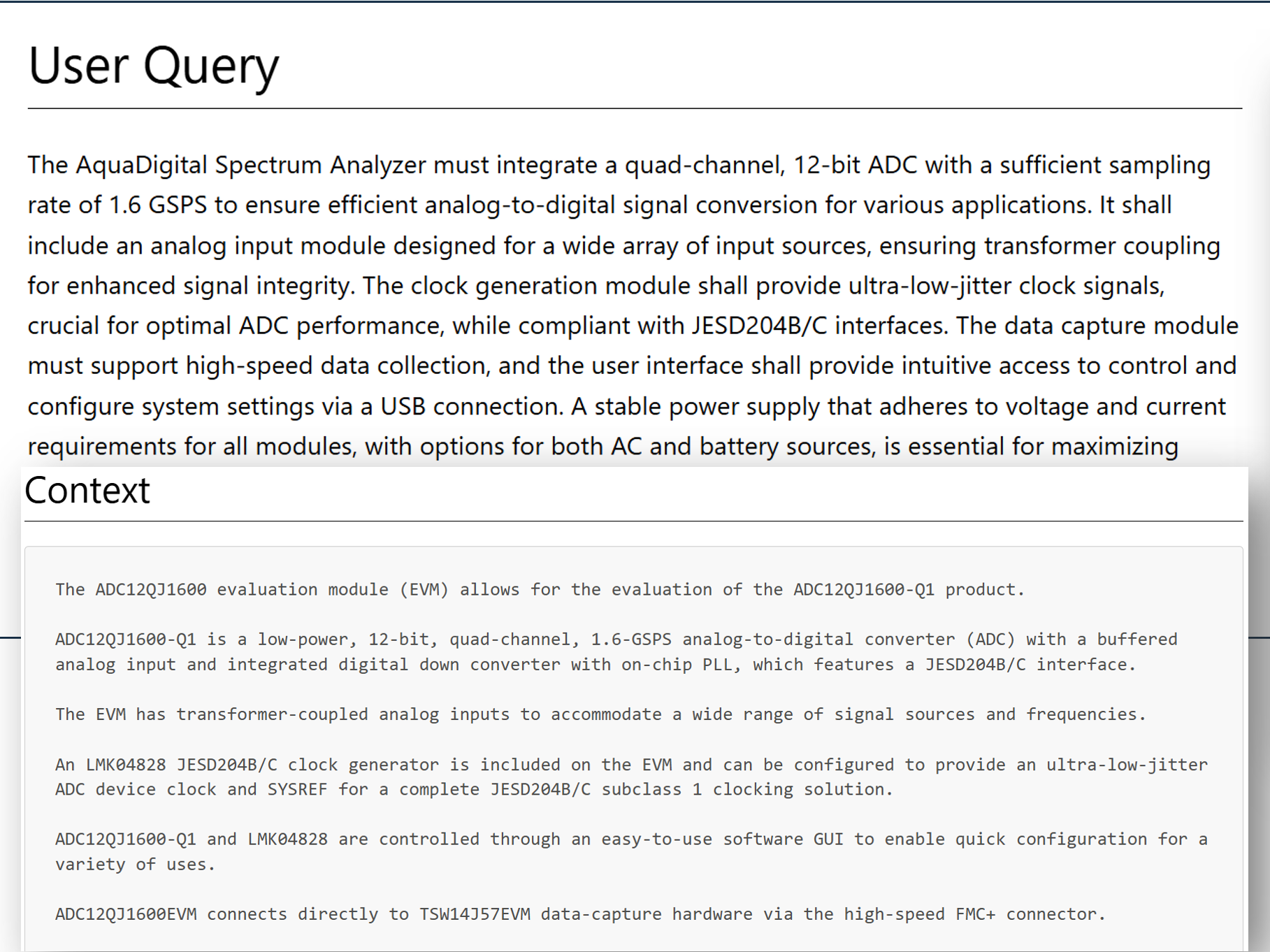}}
    \caption{Augmented Prompt: User Query and Context.}
    \label{fig:prompt}
\end{subfigure}
\hfill
\begin{subfigure}[t]{0.48\textwidth}
    \centering
    \fbox{\includegraphics[width=\linewidth]{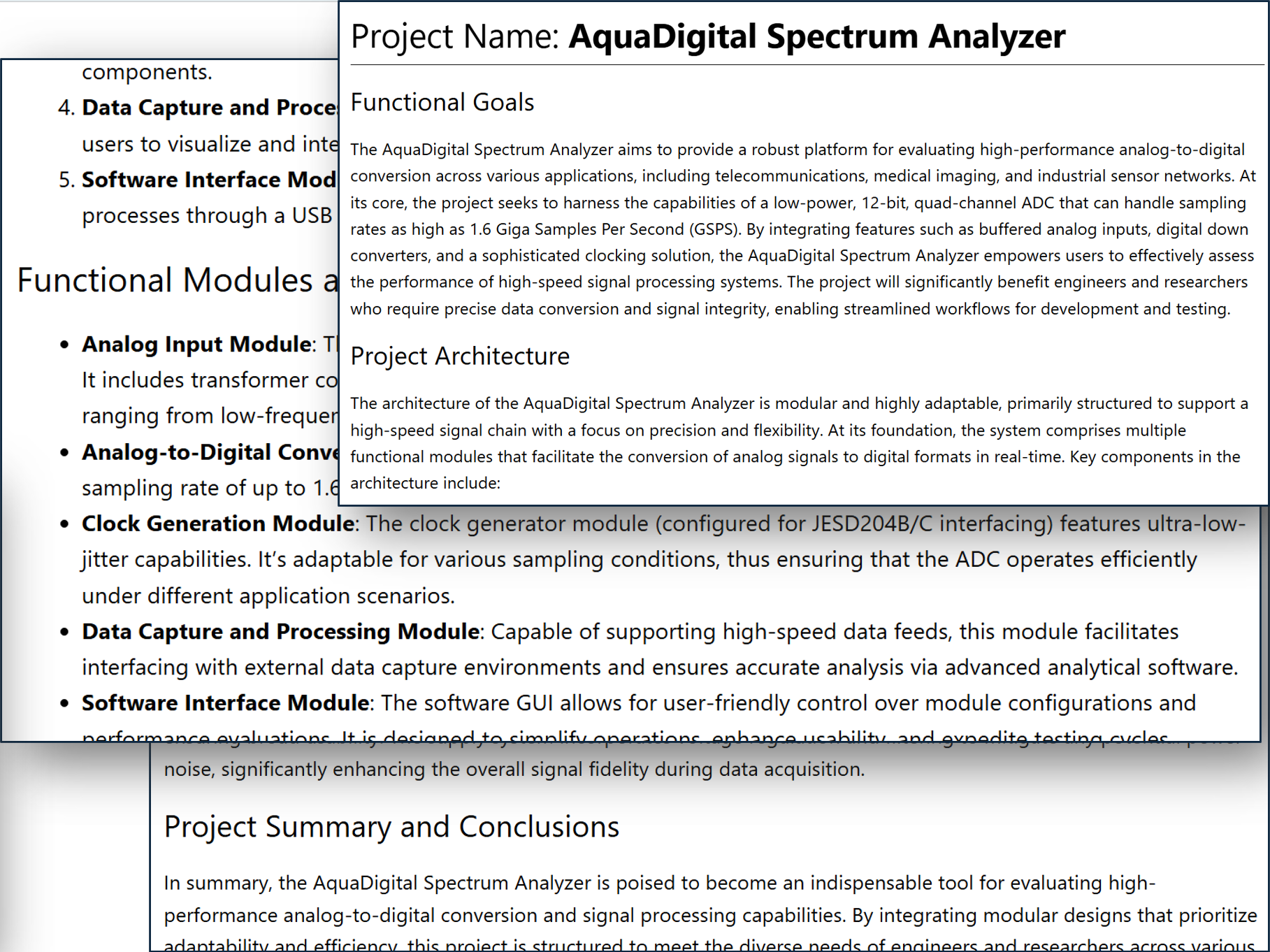}}
    \caption{Completion: Reference (project description).}
    \label{fig:completion}
\end{subfigure}
\caption{One dataset example for illustrating the use case. We prompt an RAG-Generator to generate b), a requirements engineering document, integrating information from a).}
\label{fig:data_example}
\end{figure*}
%
%
The dataset models a requirements engineering workflow in the electronics domain, targeting an electronic design automation setting in which an LLM assistant responds to brief project descriptions from users with full, structured requirement descriptions using RAG (\autoref{fig:data_example}). 
This task is motivated by the observation that electronic systems are frequently underspecified at project start (often due to gaps in technical knowledge, limited engineering experience, or time pressure) which drives costly prototype iterations, since changes made late in hardware development lead to disproportionately high costs. 
By drafting more complete requirements upfront and reusing the recurring structures common to projects within a given subdomain (e.g., IoT, power supply, or industrial sensing), such an assistant aims to reduce time-to-market and overall project cost.
%
The data originates from two sources: real manufacturer descriptions and synthetic texts derived from them.
We first extract around 14\,K textual descriptions from publicly available reference designs on the web.
Using these descriptions as truth seeds, we prompt GPT-4o to generate manufacturer-independent, synthetic project descriptions (\textit{references}) that include functionalities and specifications of the underlying text (see \autoref{fig:data_pipeline} in the Appendix).
Each reference is then summarized to simulate underspecified \textit{user queries}, deliberately excluding details.
%
Each user-query-reference pair is then combined with the original seed used for synthesis, yielding a triple per datapoint: A \textit{user query}, its \textit{reference} answer and closely related contextual information (\textit{context}).
All three entries encode the same underlying technical facts but differ in granularity, detail, and phrasing.
The final dataset comprises $\sim$12K entries (avg. 170 tokens per query, 848 per reference, 228 per context) (see \autoref{fig:data_example}).
We construct two RAG-SFT variants differing only in context: LoRA-base and LoRA-irr (inspired by \citet{bhushan-etal-2025-parag} and \citet{yoranmaking2024}).
We define \emph{relevant context} as the truth seed used for data synthesis, which we split into up to three chunks of 128 tokens each, while \emph{irrelevant context} refers to three randomly selected chunks.
LoRA-base adds to each query the relevant context and LoRA-irr duplicates each query--reference pair with two context variants (relevant and irrelevant).
\subsection{Finetuning}
We select \texttt{Qwen2.5-7B-Instruct}\cite{Yang2024Qwen25TR} and \texttt{Olmo-3-7B-Instruct}\cite{olmo2025olmo} as the base models and employ supervised fine-tuning, using LoRA\,\cite{hu2022lora} (Hyperparameters: $r = 16$, $\alpha = 32$, batch size $=32$, constant learning rate $=4\times10^{-4}$, drop out $=0.2$).
We train for two epochs on the LoRA-base and for one epoch on the LoRA-irr dataset: LoRA-irr contains two versions of every prompt, a single epoch effectively exposes each underlying sample twice. 

\subsection{Evaluation}

We use a unified test set of 2\,K instances, created by applying the two context conditions---relevant and irrelevant---to the same 1\,K test prompts, that were held out from the training process. 
This allows a controlled comparison across adapted models, including ablations (e.g., LoRA-base vs. LoRA-irr) and behaviour in unseen scenarios (e.g., LoRA-base on irrelevant context).

\paragraph{Inference}
For each test prompt, we generate one completion using greedy decoding rather than sampling, to best reflect the adapted token distribution.

\paragraph{Standard NLP metrics}
We compare generated and reference responses using Rouge \cite{lin-2004-rouge} and BertScore \cite{bert-score}.

\paragraph{Claim-based evaluation pipeline}
The first step in \textsc{C-FEX} (Claim Extraction + Fact Evaluation) (\autoref{fig:placeholder}) is to extract claims from the generated response and its corresponding source texts, the augmented prompt and reference response for each example, using an LLM. 
For each generated response, we deduplicate the generated claims, using dense embeddings and cosine similarity (threshold: 0.95). 
Then, for each response claim, we compute embeddings, and retrieve a fixed top-$5$ set of candidate evidence claims from each source, via cosine similarity. 
An LLM-as-a-judge then verifies support by attributing each generated claim against its candidate subset and records if the claim is supported by one of the three sources. 
We further batch grounding judgments by evaluating eight generated claims jointly per LLM call, each against its candidate set, yielding $8 \times 15$ claim-evidence comparisons per call and substantially reducing LLM invocations.
For all embedding-based steps, we use \texttt{Qwen3-Embedding-0.6B} as the default model, and for the llm-based steps \texttt{Qwen3-Next-80B-A3B-Instruct}. 
Due to the high computational cost of LLM-based pipelines, we evaluate only a subsample of 512 examples or 50\% per test condition, rather than the full test set.

\paragraph{Claim-based metrics}\label{sec:metrics}

Let $\hat{\mathcal{C}}_{\mathcal{G}}$, $\mathcal{C}_{\mathcal{R}}$, 
$\mathcal{C}_{\mathcal{C}}$, and $\mathcal{C}_{\mathcal{U}}$ 
denote \textit{claims} from the generated response, reference, 
retrieved context, and user query. 
We define
$\mathcal{S}_{\mathcal{R}} \subseteq \hat{\mathcal{C}}_{\mathcal{G}}$,
$\mathcal{S}_{\mathcal{C}} \subseteq \hat{\mathcal{C}}_{\mathcal{G}}$, and
$\mathcal{S}_{\mathcal{U}} \subseteq \hat{\mathcal{C}}_{\mathcal{G}}$
as the subsets of generated claims judged by the LLM to be supported by the reference, context, and user query, respectively. We report the following metrics:

\begin{itemize}
\item \textbf{Support Rate} ($\text{Prec}_{\text{any}}$) Proportion of generated claims supported by at least one source:
\begin{equation*}
\text{Prec}_{\text{any}}
=
\frac{|\mathcal{S}_{\mathcal{R}} \cup \mathcal{S}_{\mathcal{C}} \cup \mathcal{S}_{\mathcal{U}}|}
     {|\hat{\mathcal{C}}_{\mathcal{G}}|}
\end{equation*} 
\item \textbf{Reference-Backed Precision and Recall} Proportion of gen./ref. claims supported by at least one ref./gen. claim
\begin{equation*}
\hspace{-1em}
\text{Prec}_{\text{ref}}
=
\frac{|\mathcal{S}_{\mathcal{R}}|}
     {|\hat{\mathcal{C}}_{\mathcal{G}}|}
\quad \text{and} \quad
\text{Rec}_{\text{ref}}
=
\frac{|\mathcal{S}_{\mathcal{R}}|}
     {|\mathcal{C}_{\mathcal{R}}|}
\end{equation*}
\item \textbf{Parametric Knowledge Precision} (\textbf{PKP}) Proportion of correct generated claims not grounded in the user query or context:
\begin{equation*}
\text{PKP}
=
\frac{|\mathcal{S}_{\mathcal{R}} \setminus (\mathcal{S}_{\mathcal{C}} \cup \mathcal{S}_{\mathcal{U}})|}
{|\hat{\mathcal{C}}_{\mathcal{G}} \setminus (\mathcal{S}_{\mathcal{C}} \cup \mathcal{S}_{\mathcal{U}})|}
\end{equation*}
\item \textbf{Parametric Rate (PR)} Proportion  of generated claims not supported by the augmented prompt:
\begin{equation*}
\text{PR}
=
\frac{\left|\hat{\mathcal{C}}_{\mathcal{G}}\setminus(\mathcal{S}_U \cup \mathcal{S}_C)\right|}
{\left|\hat{\mathcal{C}}_{\mathcal{G}}\right|}
\end{equation*}
\item \textbf{Self-Knowledge} (\textbf{SK}) Proportion of correct generated claims not supported by either retrieved context or the user query:
\begin{equation*}
\text{SK}
=
\frac{|\mathcal{S}_{\mathcal{R}} \setminus (\mathcal{S}_{\mathcal{C}} \cup \mathcal{S}_{\mathcal{U}})|}
     {|\hat{\mathcal{C}}_{\mathcal{G}}|}
\end{equation*}
\end{itemize}
%
SK follows prior work from \citet{ru2024ragchecker} and, as it is normalized by the total number of generated claims $|\hat{\mathcal{C}}_{\mathcal{G}}|$, including those attributable to context or user query, it combines parametric quality with quantity.
In contrast, PKP conditions explicitly on the subset of claims that \textit{require} parametric knowledge, since they do not appear in the augmented prompt: parametric reference facts and hallucinations ($|\hat{\mathcal{C}}_{\mathcal{G}} \setminus (\mathcal{S}_{\mathcal{C}} \cup \mathcal{S}_{\mathcal{U}})|$ = $|\mathcal{P}_{\textit{ref}}|$ + $|\mathcal{H}|$ in \autoref{fig:placeholder}).
Therefore, we decompose $\text{SK} = \text{PKP} \times \text{PR}$; with PR measuring the proportion of generated claims that require parametric knowledge, and PKP measuring the correctness of those claims with respect to the reference. 


\begin{table*}[t]
\centering
\small
\setlength{\tabcolsep}{4pt}
\begin{subtable}{\textwidth}
\caption{Testprompts: Relevant context chunk(s) only.}
\centering
\begin{adjustbox}{width=\textwidth}
\begin{tabular}{l | ccccc | ccc}
\toprule
Model
 & R-1 & R-2 & R-L & BS-Prec & BS-Rec
 & Prec$_\text{ref}$ & Prec$_\text{any}$ & Rec$_\text{ref}$ \\
\midrule
Qwen2.5-7B
 & 50.81 & 21.01 & 25.38 & 77.63 & 76.12
 & 63.49 & 73.26 & 79.64 \\
\hspace{0.05cm}\,+ LoRA-base
 & \textbf{66.02}\,{\scriptsize($+29.94\%$)} & \textbf{30.13}\,{\scriptsize($+43.41\%$)} & \textbf{33.76}\,{\scriptsize($+33.02\%$)}
 & 82.50\,{\scriptsize($+6.27\%$)} & \textbf{79.29}\,{\scriptsize($+4.16\%$)}
 & 81.01\,{\scriptsize($+27.59\%$)} & 89.44\,{\scriptsize($+22.09\%$)} & 81.14\,{\scriptsize($+1.88\%$)} \\
\hspace{0.05cm}\,+ LoRA-irr
 & 65.81\,{\scriptsize($+29.52\%$)} & \underline{29.94}\,{\scriptsize($+42.50\%$)} & \underline{33.44}\,{\scriptsize($+31.76\%$)}
 & 82.52\,{\scriptsize($+6.30\%$)} & 79.16\,{\scriptsize($+3.99\%$)}
 & 81.24\,{\scriptsize($+27.96\%$)} & 89.46\,{\scriptsize($+22.11\%$)} & \underline{81.75}\,{\scriptsize($+2.65\%$)} \\
\midrule
Olmo-3-7B
 & 48.23 & 17.56 & 20.94 & 72.18 & 78.36
 & 61.25 & 69.13 & 76.91 \\
\hspace{0.05cm}\,+ LoRA-base
 & 65.77\,{\scriptsize($+36.37\%$)} & 29.92\,{\scriptsize($+70.39\%$)} & 33.43\,{\scriptsize($+59.65\%$)}
 & \textbf{82.58}\,{\scriptsize($+14.41\%$)} & \underline{79.22}\,{\scriptsize($+1.10\%$)}
 & \underline{81.36}\,{\scriptsize($+32.83\%$)} & \underline{89.57}\,{\scriptsize($+29.57\%$)} & 81.27\,{\scriptsize($+5.67\%$)} \\
\hspace{0.05cm}\,+ LoRA-irr
 & \underline{65.84}\,{\scriptsize($+36.51\%$)} & 29.92\,{\scriptsize($+70.39\%$)} & \underline{33.44}\,{\scriptsize($+59.69\%$)}
 & \underline{82.55}\,{\scriptsize($+14.37\%$)} & 78.99\,{\scriptsize($+0.80\%$)}
 & \textbf{81.54}\,{\scriptsize($+33.13\%$)} & \textbf{89.65}\,{\scriptsize($+29.68\%$)} & 80.10\,{\scriptsize($+4.15\%$)} \\
\midrule
Qwen2.5-72B
 & 51.23 & 21.34 & 25.08 & 77.98 & 77.22
 & 60.18 & 68.68 & \textbf{81.84} \\
\bottomrule
\end{tabular}
\end{adjustbox}
\label{tab:sft-relevant}
\end{subtable}
\vspace{0.5cm}
\begin{subtable}{\textwidth}
\caption{Testprompts: Irrelevant context chunk(s) only.}
\centering
\begin{adjustbox}{width=\textwidth}
\begin{tabular}{l | ccccc | ccc}
\toprule
Model
 & R-1 & R-2 & R-L & BS-Prec & BS-Rec
 & Prec$_\text{ref}$ & Prec$_\text{any}$ & Rec$_\text{ref}$ \\
\midrule
Qwen2.5-7B
 & 50.46 & 20.33 & 24.98 & 77.09 & 75.13
 & 59.16 & 67.01 & 75.35 \\
\hspace{0.05cm}\,+ LoRA-base
 & 54.27\,{\scriptsize($+7.55\%$)} & 19.73\,{\scriptsize($-2.95\%$)} & 25.59\,{\scriptsize($+2.44\%$)}
 & 73.47\,{\scriptsize($-4.70\%$)} & 71.86\,{\scriptsize($-4.35\%$)}
 & 39.92\,{\scriptsize($-32.52\%$)} & 44.07\,{\scriptsize($-34.23\%$)} & 44.97\,{\scriptsize($-40.32\%$)} \\
\hspace{0.05cm}\,+ LoRA-irr
 & \underline{64.01}\,{\scriptsize($+26.85\%$)} & \underline{28.30}\,{\scriptsize($+39.20\%$)} & \underline{32.11}\,{\scriptsize($+28.54\%$)}
 & \underline{82.01}\,{\scriptsize($+6.38\%$)} & \textbf{77.87}\,{\scriptsize($+3.65\%$)}
 & \textbf{79.31}\,{\scriptsize($+34.06\%$)} & \textbf{85.97}\,{\scriptsize($+28.29\%$)} & \underline{76.15}\,{\scriptsize($+1.06\%$)} \\
\midrule
Olmo-3-7B
 & 46.73 & 16.29 & 20.02 & 71.02 & 77.28
 & 54.58 & 59.54 & 69.87 \\
\hspace{0.05cm}\,+ LoRA-base
 & 59.48\,{\scriptsize($+27.28\%$)} & 24.20\,{\scriptsize($+48.56\%$)} & 28.81\,{\scriptsize($+43.91\%$)}
 & 77.82\,{\scriptsize($+9.57\%$)} & 75.32\,{\scriptsize($-2.54\%$)}
 & 57.51\,{\scriptsize($+5.37\%$)} & 62.67\,{\scriptsize($+5.26\%$)} & 64.59\,{\scriptsize($-7.56\%$)} \\
\hspace{0.05cm}\,+ LoRA-irr
 & \textbf{64.17}\,{\scriptsize($+37.32\%$)} & \textbf{28.35}\,{\scriptsize($+74.03\%$)} & \textbf{32.19}\,{\scriptsize($+60.79\%$)}
 & \textbf{82.03}\,{\scriptsize($+15.50\%$)} & \underline{77.70}\,{\scriptsize($+0.54\%$)}
 & \underline{78.86}\,{\scriptsize($+44.49\%$)} & \underline{85.59}\,{\scriptsize($+43.75\%$)} & 74.87\,{\scriptsize($+7.16\%$)} \\
\midrule
Qwen2.5-72B
 & 50.18 & 20.26 & 24.34 & 77.26 & 76.31
 & 53.56 & 59.30 & \textbf{77.77} \\
\bottomrule
\end{tabular}
\end{adjustbox}
\label{tab:sft-irrelevant}
\end{subtable}
\caption{Absolute scores (in percentage points) with relative finetuning gain of RAG-SFT in parentheses. We repeat the LLM-Judgment three times, compute the mean per-example standard deviation across those three runs and aggregate over all models and test cases: Prec$_{ref}$$\pm1.01$, Prec$_{any}$$\pm0.75$ and Rec$_{ref}$$\pm0.76$ (in pp).} 
\label{tab:sft-combined}
\end{table*}

\paragraph{Evaluation of LLM pipeline steps}
We carry out a small human evaluation, judging the llm-based outputs using Claim-Extraction-Precision and Attribution-Accuracy.
Claim-Extraction-Precision is assessed by presenting $100$ claims and their extraction sources to a human judge, who decides whether each claim was faithfully extracted from its source.
For attribution evaluation $100$ response claims and their retrieved candidates (top-$15$) are presented to let the human judge whether each candidate claim is a source of the response claim: E.g., is the candidate claim, extracted from the context section a source for the response claim? $\rightarrow$ Label: (not) supported by context. 
Attribution-Accuracy is then computed by comparing the LLM-as-a-Judge labels against the human judgments.
To assess Inter-Annotator-Agreement, we employ another person annotating the same samples and compute Cohens-Kappa and per-label average agreement.
We report Kappa-Scores of 66.20\% (avg. agr. 99.00\%) for claim extraction and 42.19\% (avg. agr. 76.33\%) for claim attribution by LLM-as-a-judge, reaching substantial and moderate agreement respectively\,\cite{kappainterpret}.
We choose randomly between both labels if the annotators disagree and report a final Claim-Extraction-Precision of 99.00\% and an Attribution-Accuracy of 80.67\%.
\section{Results and Discussion}
We report all metrics before and after fine-tuning in \autoref{tab:sft-combined} and the metrics for evaluating parametric knowledge in \autoref{tab:pk-combined} and organize the discussion around our three research questions.
\subsection{Does RAG-SFT transfer to requirements-document generation?}

RAG-SFT works well for our task, with both 7B models closing the gap to a model an order of magnitude larger.

Under relevant context, LoRA-irr's contrastive training improves both base models on almost every quality dimension we measure (\autoref{tab:sft-relevant}), validating the approach of \citet{bhushan-etal-2025-parag} and \citet{yoranmaking2024} in a setting well outside of document Q\&A on which they were developed. 
The fine-tuned 7B models surpass the 72B vanilla-RAG baseline on every quality dimension except Rec$_{ref}$, where the 10$\times$ additional parameters help covering slightly more reference knowledge at test time.
The two training strategies diverge only when retrieval fails. LoRA-base is comparable to LoRA-irr under relevant context, but collapses when tested out-of-distribution on irrelevant chunks (\autoref{tab:sft-irrelevant}). 
For Qwen-7B, Prec$_{ref}$ and Rec$_{ref}$ fall by 32.52\% and 40.32\% relative, whereas LoRA-irr remains stable across both conditions. 
This collapse is due to training the model to overly adhere to the context and blindly integrating it in the answer. 
When we consider the fraction of claims attributed to the context section, we report that it increased after LoRA-base fine-tuning from 5.2\% to 32.6\% for Qwen-7B and from 6.5\% to 16.1\% for Olmo-7B.
The two figures also explain why the collapse is sharper for Qwen than for Olmo: the model that comes to lean hardest on the context is the one that suffers most when the context is uninformative. 
Contrastive exposure to irrelevant chunks (LoRA-irr) prevents this failure mode while retaining the in-distribution gains. 
\subsection{Do standard NLP metrics agree with claim-based factuality?}
\begin{figure*}[ht]
\centering
\begin{subfigure}{0.48\linewidth}
    \centering
    \includegraphics[width=\linewidth]{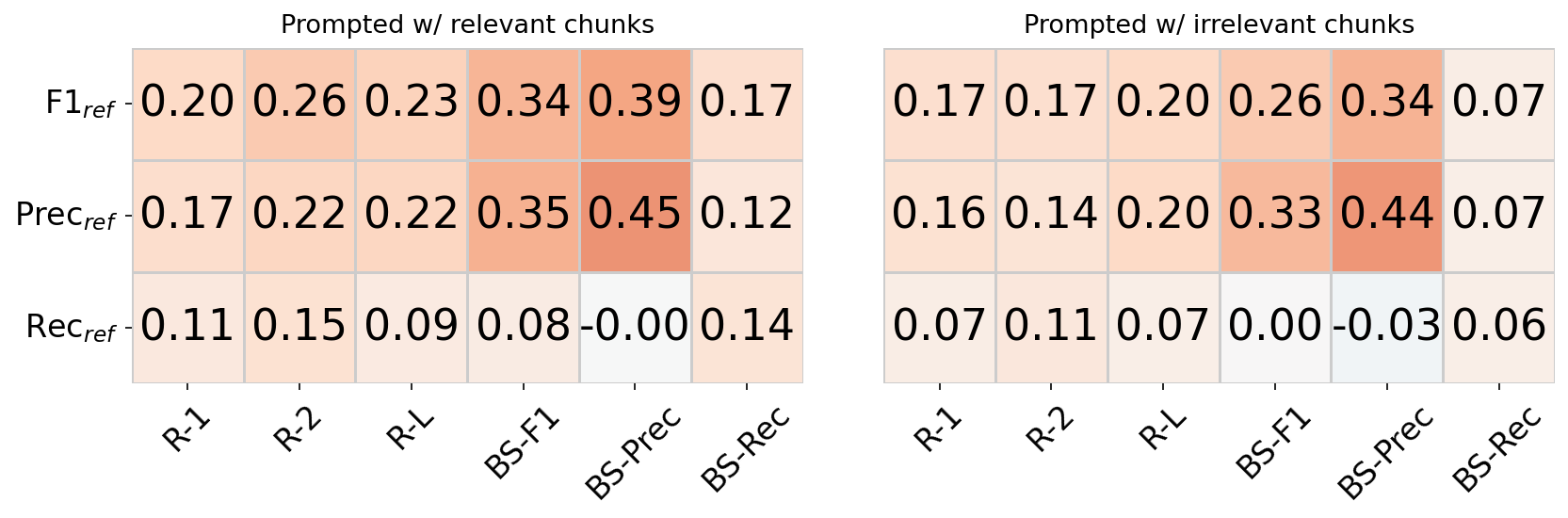}
    \caption{Qwen2.5-7B}
    \label{fig:correlate-7B-base}
\end{subfigure}
\hfill
\begin{subfigure}{0.48\linewidth}
    \centering
    \includegraphics[width=\linewidth]{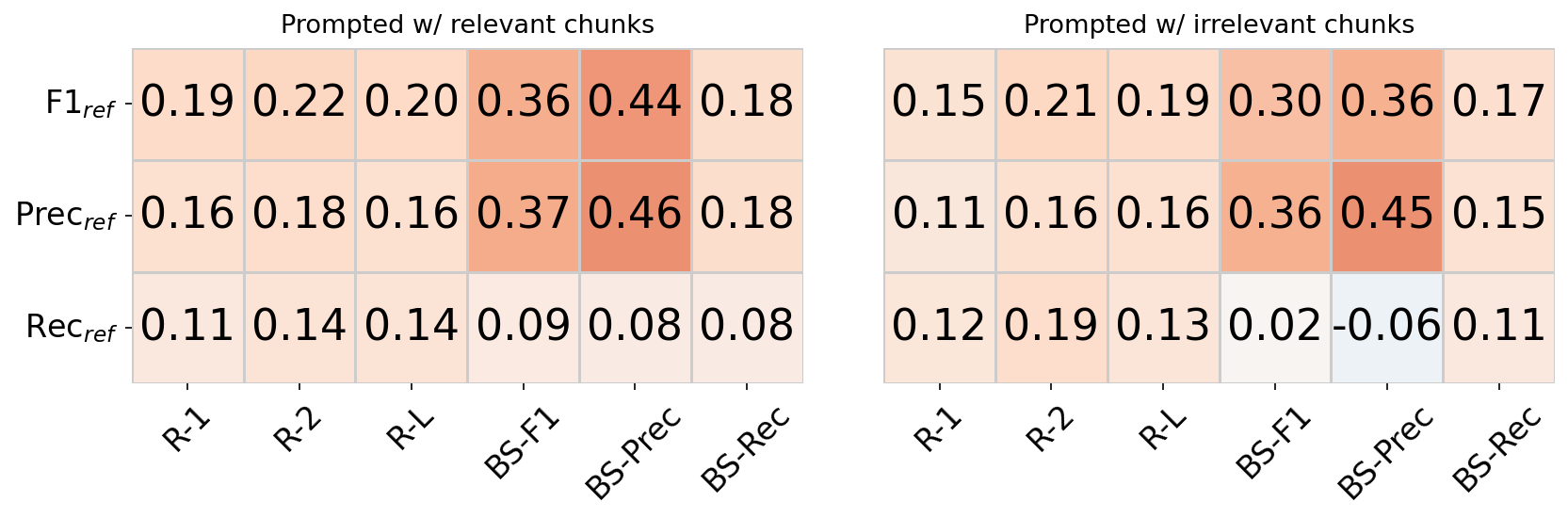}
    \caption{Qwen2.5-72B}
    \label{fig:correlate-72B-base}
\end{subfigure}

\vspace{\baselineskip}

\begin{subfigure}{0.48\linewidth}
    \centering
    \includegraphics[width=\linewidth]{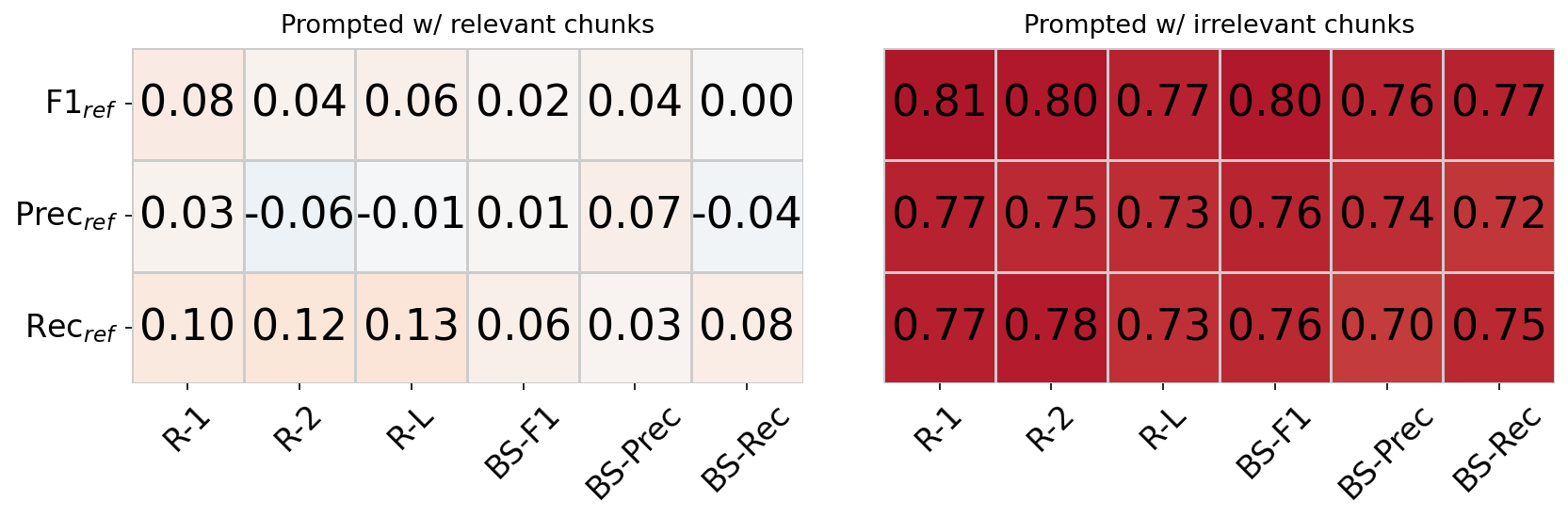}
    \caption{Qwen2.5-7B + LoRA-base}
    \label{fig:correlate-7B-lora-ki}
\end{subfigure}
\hfill
\begin{subfigure}{0.48\linewidth}
    \centering
    \includegraphics[width=\linewidth]{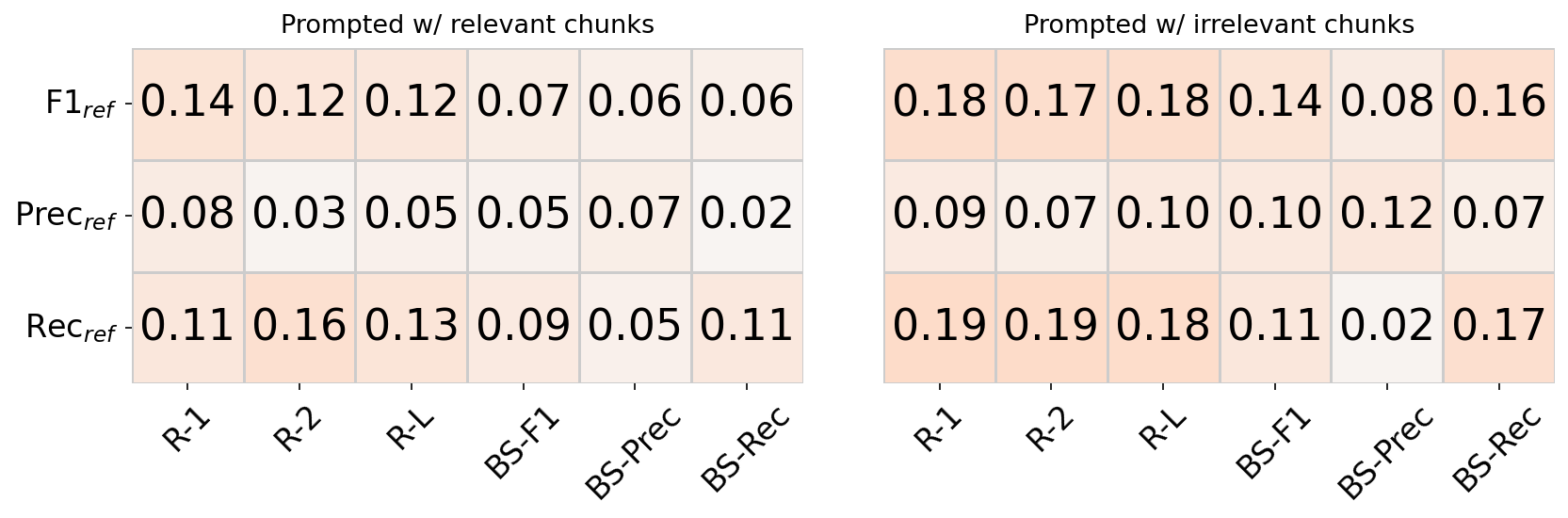}
    \caption{Qwen2.5-7B + LoRA-irr}
    \label{fig:correlate-7B-lora-parag}
\end{subfigure}
\caption{Spearman Rank Correlation of Rouge (R), BertScore (BS) and claim-based metrics.}
\label{fig:correlations}
\end{figure*}
Rouge and BertScore capture claim-based factuality rather poorly, and can mislead about both the size and the direction of fine-tuning effects.

A first structural observation is that the values of the claim-based metrics are not compressed into such a narrow band like Rouge and BertScore, which is a known limitation of these standard metrics (see \autoref{tab:sft-combined}).\footnote{\citet{bert-score} proposed rescaling BertScore by an empirical lower bound (e.g. scores of randomly paired sentences from a corpus), but this is costly for long-context embedding models, which have no precomputed rescaling factors.}
%
If we look at fine-tuning gains, we see that the magnitude of the gain is very different for each metric group, but they mostly agree on the sign---one outlier is Rouge for the LoRA-base model on irrelevant chunks (\autoref{tab:sft-irrelevant}): On both model architectures, Qwen2.5 and Olmo-3, Rouge and the other recall metrics disagree on the fine-tuning result---Rouge states a positive gain, and, BertScore-Recall and Rec$_{ref}$ state a decrease. 
This is invisible to Rouge, likely because the responses remain lexically similar to the learned references in large parts, concealing hallucinations behind plausibly formulated text. To illustrate this, we show one example of Qwen-7B in the Appendix (\autoref{fig:phen_rouge}), where Rouge-1 and Rec$_{ref}$ are disagreeing (Rouge-1 $=61.64\%$ vs. Rec$_{ref}=7.14\%$), revealing that the prediction is largely overlapping on words and structure with the reference, but contains a lot of technical values, which are not supported by the reference. 
%

To quantify the agreement, we compute spearman rank correlation $r$ of Rouge and BertScore vs. our claim-based metrics (\autoref{fig:correlations}, for Qwen2.5-7\,/\,72B).
Correlations are generally low, confirming that Rouge and BertScore are, if at all, a weak proxy measure for reference-backed factuality, but two exceptions stand out:
First, BertScore-Precision correlates moderately with Prec$_{ref}$ on the base models ($r=0.44$--$0.45$), a useful finding since BertScore is far cheaper than our llm-based pipeline and may serve as a lightweight precision proxy on this kind of technical text. 
Second, the LoRA-base model tested out-of-distribution shows unusually high correlations ($r=0.72$--$0.81$); we assume due to the model collapse the failures in the responses are so broad that all metrics, more or less coincidentally agree. 
Additionally, we see correlations diminish after training in three of the four cases (the collapse being the sole exception); because RAG-SFT lead to an general increase in metric scores, genuine bad answers become rare, and the coincidental agreement that low-quality outputs may produce disappears, leaving the metrics increasingly orthogonal.
The same patterns hold, more weakly, for Olmo-3 (see Appendix \autoref{fig:correlations-olmo}). 
\subsection{How does fine-tuning change parametric knowledge?}
\begin{table}[ht]
\centering
\small
\begin{subtable}{\linewidth}
\caption{Testprompts: Relevant context chunk(s) only.}
\centering
\begin{adjustbox}{width=\linewidth}
\begin{tabular}{l ccccc}
\toprule
Model & PKP & PR & SK & $|\mathcal{P}_{ref}|$* & $|\mathcal{H}|$* \\
\midrule
\multicolumn{4}{l}{\textit{Base}} \\
Qwen2.5-7B  & 32.59 & 39.70 & 12.93 & 3.8 & 7.9 \\
Olmo-3-7B    & 32.39 & 45.70 & 14.79 & 6.2 & 12.9 \\
Qwen2.5-72B & 30.39 & 45.00 & 13.68 & 4.4 & 10.1 \\
\midrule
\multicolumn{4}{l}{\textit{LoRA-base}} \\
Qwen2.5-7B  & 62.70 & 28.30 & 17.75 & 3.7 & 2.2 \\
Olmo-3-7B    & 62.55 & 27.80 & 17.42 & 3.7 & 2.2 \\
\midrule
\multicolumn{4}{l}{\textit{LoRA-irr}} \\
Qwen2.5-7B  & 62.70 & 28.30 & 17.71 & 3.8 & 2.3 \\
Olmo-3-7B    & 63.17 & 28.10 & 17.76 & 3.7 & 2.1 \\
\bottomrule
\end{tabular}
\end{adjustbox}
\label{tab:pk-relevant}
\end{subtable}
\vspace{0.5cm}
\begin{subtable}{\linewidth}
\caption{Testprompts: Irrelevant context chunk(s) only.}
\centering
\begin{adjustbox}{width=\linewidth}
\begin{tabular}{l ccccc}
\toprule
Model & PKP & PR & SK & $|\mathcal{P}_{ref}|$* & $|\mathcal{H}|$* \\
\midrule
\multicolumn{4}{l}{\textit{Base}} \\
Qwen2.5-7B  & 32.07 & 41.60 & 13.35 & 4.0 & 8.5 \\
Olmo-3-7B   & 30.64 & 50.20 & 15.39 & 7.2 & 16.3 \\
Qwen2.5-72B & 27.93 & 51.70 & 14.43 & 5.0 & 13.0 \\
\midrule
\multicolumn{4}{l}{\textit{LoRA-base}} \\
Qwen2.5-7B  & 28.56 & 34.80 & 9.94 & 2.2 & 5.5 \\
Olmo-3-7B    & 39.40 & 36.70 & 14.47 & 3.2 & 5.0 \\
\midrule
\multicolumn{4}{l}{\textit{LoRA-irr}} \\
Qwen2.5-7B  & 58.04 & 32.00 & 18.60 & 3.8 & 2.8 \\
Olmo-3-7B    & 57.26 & 32.20 & 18.46 & 3.7 & 2.8 \\
\bottomrule
\end{tabular}
\end{adjustbox}
\label{tab:pk-irrelevant}
\end{subtable}
\caption{Absolute scores of the parametric knowledge metrics (\,*\,= mean claims per example). Standard deviations across three LLM-Judgments, all models and test cases: PKP$\pm1.89$, PR$\pm1.09$ and SK$\pm0.87$ (in pp).}
\label{tab:pk-combined}
\end{table}
Fine-tuning does not teach the models new correct parametric knowledge; it trades the quantity of parametric output for its reliability. 
We separate these two aspects using the decomposition SK = PKP $\times$ PR, and they move in opposite directions.

We see in \autoref{tab:pk-combined}, that the parametric output (PR) of the base models, under both test scenarios, is clearly higher, than that of the finetuned ones---both RAG-SFT regimes lead to less parametric output, or, conversely, to stronger adherence to the augmented prompt ($1-$PR); the second is a plausible training outcome, as shown by related work.\footnote{E.g. \citet{lin2025resisting} show this tendency, framing standard RAG and GRPO-with-RAG as over-focusing on contextual reasoning at the expense of parametric knowledge.}
When retrieval fails, all models shift back toward their weights (PR in \autoref{tab:pk-irrelevant} $>$ \autoref{tab:pk-relevant}), with a larger shift for the fine-tuned models: they have more prompt-adherence to fall back from.
The quality of that parametric output (PKP) moves in the opposite direction.
The base models cluster around a PKP level of $31.00\%$ on average ($27.93\%$--$32.59\%$) vs. $54.30\%$ of the trained ones ($28.56\%$--$63.17\%$). 
This quality difference is mainly driven by much less hallucinations in the PKP denominator of the trained models in comparison to the base ones and analyzing the effect per response makes it clear: on average the base models produce $5.1$ correct parametric claims and the trained ones $3.475$ ($|\mathcal{P}_{ref}|$* in \autoref{tab:pk-combined}), a $31.86\%$ decrease, whereas incorrect parametric claims are cut from $11.45$ to $2.4$ per answer, a $79.04\%$ decrease ($|\mathcal{H}|$* in \autoref{tab:pk-combined}). 
So in fact, after training the models do not output \textit{more} correct claims from its weights, they output (slightly) \textit{less}, but \textit{mostly} refrain from outputting incorrect ones.
This is precisely the shift that Self-Knowledge (SK) overlooks. 
Because SK keeps prompt-supported claims in its denominator, it entangles the quality and the quantity of parametric knowledge, and when those two factors move in opposition---less frequent but more reliable parametric output (PR$\downarrow$, PKP$\uparrow$), or the reverse---they cancel, leaving SK nearly unchanged.
To a certain extent, we observe exactly this: averaged before and after RAG-SFT, SK moves only approx. $3.48\%$, a rather small change comparable to the scale of change of its underlying factors.
The same masking appears more clearly when probing retrieval robustness (\autoref{tab:pk-relevant} vs. \autoref{tab:pk-irrelevant}): both LoRA-irr models report SK of $17.71\% / 17.76\%$ under successful retrieval versus $18.60\% / 18.46\%$ under failed retrieval, an apparent non-effect, if we look at the observed standard deviation of $\pm0.87\%$ of SK, that hides the fact that PKP and PR are each moving in opposite directions.

\section{Conclusions}
We studied supervised RAG fine-tuning for long-form requirements-document generation in the electronics domain, a setting that departs from the question answering tasks where RAG-SFT is usually evaluated. 
To measure reference-backed factuality on these long technical responses, where Rouge and BertScore are weak proxies, we introduced \textsc{C-FEX}, a claim-based evaluation pipeline that attributes every response claim to its origin, and PKP, which isolates the correctness of claims drawn from the model's weights. 
Our experiments show that 7B adapters match or exceed a 72B baseline and, with contrastive relevant/irrelevant training, remain robust when retrieval fails. 
Crucially, the claim-based decomposition reveals a shift that aggregated metrics obscure: fine-tuning does not teach the models new correct parametric knowledge---they emit fewer correct parametric claims (-31.86\%)---but it sharply suppresses hallucinations (-79.04\%), trading the quantity of parametric output for its reliability. 
Because Self-Knowledge multiplies these two opposing factors (SK = PKP $\times$ PR), it stays nearly flat and hides this change, whereas reporting PR and PKP separately makes it visible. 
We therefore suggest claim-based evaluation for long-form technical generation, and treating the rate and the quality of parametric output as distinct quantities rather than collapsing them into one score.

\newpage
\section*{Limitations}

Because the scope of this study was parametric knowledge, we report "context adherence" only as a byproduct: $1-$PR clearly shows that RAG-SFT led to greater attention to information in the augmented prompt, but we did not report a finer analysis. 
We therefore add the complete source-attribution results to the Appendix, to show the different effects fine-tuning had on the two sections of the augmented prompt, namely the retrieved context and the user query (see \autoref{fig:source-distri}).

We further acknowledge that our definition of parametric knowledge may label externally valid claims---e.g. a technical specification not originating from the augmented prompt or reference document still might be valid if judged by a domain expert---in the model responses as hallucinations. 
Our definition is a \textit{lower bound}: it excludes any claim that is (i) absent from the reference response and (ii) overlapping with a claim in the augmented prompt that itself overlaps with the reference response. 

Additionally, the LLM-as-a-Judge claim attribution achieves only $80.67\%$ accuracy relative to human judgments, introducing evaluation noise---without further error analysis, we cannot determine how this affects our results. 
Nevertheless, we argue that we have not overclaimed any finding, since our conclusions rest on rather strong shifts in the metrics and were partly observed in related work, albeit on different datasets. 

\section*{Ethical Considerations}
Using an LLM-based pipeline involves a trade-off between cost (monetary and ecological footprint) and accuracy, as larger models or commercial APIs are generally better suited for off-the-shelf usage and therefore produce less noisy results. 
This dilemma of allocating more resources and raising the ecological footprint, without certainty about the clarity of the resulting signals, should not be neglected when using this system.   
\bibliography{custom}

@inproceedings{bhushan-etal-2025-parag,
    title = "Systematic Knowledge Injection into Large Language Models via Diverse Augmentation for Domain-Specific {RAG}",
    author = "Bhushan, Kushagra  and
      Nandwani, Yatin  and
      Khandelwal, Dinesh  and
      Gupta, Sonam  and
      Pandey, Gaurav  and
      Raghu, Dinesh  and
      Joshi, Sachindra",
    editor = "Chiruzzo, Luis  and
      Ritter, Alan  and
      Wang, Lu",
    booktitle = "Findings of the Association for Computational Linguistics: NAACL 2025",
    month = apr,
    year = "2025",
    address = "Albuquerque, New Mexico",
    publisher = "Association for Computational Linguistics",
    url = "https://aclanthology.org/2025.findings-naacl.329/",
    doi = "10.18653/v1/2025.findings-naacl.329",
    pages = "5922--5943",
    ISBN = "979-8-89176-195-7"
}

@inproceedings{
zhang2024raft,
title={{RAFT}: Adapting Language Model to Domain Specific {RAG}},
author={Tianjun Zhang and Shishir G Patil and Naman Jain and Sheng Shen and Matei Zaharia and Ion Stoica and Joseph E. Gonzalez},
booktitle={First Conference on Language Modeling},
year={2024},
url={https://openreview.net/forum?id=rzQGHXNReU}
}

@inproceedings{wang-etal-2024-searching,
    title = "Searching for Best Practices in Retrieval-Augmented Generation",
    author = "Wang, Xiaohua  and
      Wang, Zhenghua  and
      Gao, Xuan  and
      Zhang, Feiran  and
      Wu, Yixin  and
      Xu, Zhibo  and
      Shi, Tianyuan  and
      Wang, Zhengyuan  and
      Li, Shizheng  and
      Qian, Qi  and
      Yin, Ruicheng  and
      Lv, Changze  and
      Zheng, Xiaoqing  and
      Huang, Xuanjing",
    editor = "Al-Onaizan, Yaser  and
      Bansal, Mohit  and
      Chen, Yun-Nung",
    booktitle = "Proceedings of the 2024 Conference on Empirical Methods in Natural Language Processing",
    month = nov,
    year = "2024",
    address = "Miami, Florida, USA",
    publisher = "Association for Computational Linguistics",
    url = "https://aclanthology.org/2024.emnlp-main.981/",
    doi = "10.18653/v1/2024.emnlp-main.981",
    pages = "17716--17736"
}

@article{liu2024chatqa,
  title={Chatqa: Surpassing gpt-4 on conversational qa and rag},
  author={Liu, Zihan and Ping, Wei and Roy, Rajarshi and Xu, Peng and Lee, Chankyu and Shoeybi, Mohammad and Catanzaro, Bryan},
  journal={Advances in Neural Information Processing Systems},
  volume={37},
  pages={15416--15459},
  year={2024}
}

@inproceedings{xuchatqa2,
  title={ChatQA 2: Bridging the Gap to Proprietary LLMs in Long Context and RAG Capabilities},
  author={Xu, Peng and Ping, Wei and Wu, Xianchao and Xu, Chejian and Liu, Zihan and Shoeybi, Mohammad and Catanzaro, Bryan},
  booktitle={The Thirteenth International Conference on Learning Representations},
  year={2025},
  url={https://openreview.net/forum?id=cPD2hU35x3}
}

@inproceedings{lin-2004-rouge,
    title = "{ROUGE}: A Package for Automatic Evaluation of Summaries",
    author = "Lin, Chin-Yew",
    booktitle = "Text Summarization Branches Out",
    month = jul,
    year = "2004",
    address = "Barcelona, Spain",
    publisher = "Association for Computational Linguistics",
    url = "https://aclanthology.org/W04-1013/",
    pages = "74--81"
}

@inproceedings{bert-score,
  title={BERTScore: Evaluating Text Generation with BERT},
  author={Zhang, Tianyi and Kishore, Varsha and Wu, Felix and Weinberger, Kilian Q and Artzi, Yoav},
  booktitle={International Conference on Learning Representations},
  year={2020},
  url={https://openreview.net/forum?id=SkeHuCVFDr}
}

@article{gao2023ragsurvey,
  title={Retrieval-augmented generation for large language models: A survey},
  author={Gao, Yunfan and Xiong, Yun and Gao, Xinyu and Jia, Kangxiang and Pan, Jinliu and Bi, Yuxi and Dai, Yixin and Sun, Jiawei and Wang, Haofen and Wang, Haofen},
  journal={arXiv preprint arXiv:2312.10997},
  year={2023}
}

@inproceedings{yoranmaking2024,
  title={Making Retrieval-Augmented Language Models Robust to Irrelevant Context},
  author={Ori Yoran and Tomer Wolfson and Ori Ram and Jonathan Berant},
  booktitle={The Twelfth International Conference on Learning Representations},
  year={2024},
  url={https://openreview.net/forum?id=ZS4m74kZpH}
}

@inproceedings{
hu2022lora,
title={Lo{RA}: Low-Rank Adaptation of Large Language Models},
author={Edward J Hu and yelong shen and Phillip Wallis and Zeyuan Allen-Zhu and Yuanzhi Li and Shean Wang and Lu Wang and Weizhu Chen},
booktitle={International Conference on Learning Representations},
year={2022},
url={https://openreview.net/forum?id=nZeVKeeFYf9}
}

@inproceedings{jiang-etal-2025-towards-better,
    title = "Towards Better Evaluation for Generated Patent Claims",
    author = "Jiang, Lekang  and
      Scherz, Pascal A.  and
      Goetz, Stefan",
    editor = "Che, Wanxiang  and
      Nabende, Joyce  and
      Shutova, Ekaterina  and
      Pilehvar, Mohammad Taher",
    booktitle = "Proceedings of the 63rd Annual Meeting of the Association for Computational Linguistics (Volume 1: Long Papers)",
    month = jul,
    year = "2025",
    address = "Vienna, Austria",
    publisher = "Association for Computational Linguistics",
    url = "https://aclanthology.org/2025.acl-long.190/",
    doi = "10.18653/v1/2025.acl-long.190",
    pages = "3775--3788",
    ISBN = "979-8-89176-251-0"
}

@inproceedings{honovich-etal-2022-true-evaluating,
    title = "{TRUE}: Re-evaluating Factual Consistency Evaluation",
    author = "Honovich, Or  and
      Aharoni, Roee  and
      Herzig, Jonathan  and
      Taitelbaum, Hagai  and
      Kukliansy, Doron  and
      Cohen, Vered  and
      Scialom, Thomas  and
      Szpektor, Idan  and
      Hassidim, Avinatan  and
      Matias, Yossi",
    editor = "Carpuat, Marine  and
      de Marneffe, Marie-Catherine  and
      Meza Ruiz, Ivan Vladimir",
    booktitle = "Proceedings of the 2022 Conference of the North American Chapter of the Association for Computational Linguistics: Human Language Technologies",
    month = jul,
    year = "2022",
    address = "Seattle, United States",
    publisher = "Association for Computational Linguistics",
    url = "https://aclanthology.org/2022.naacl-main.287/",
    doi = "10.18653/v1/2022.naacl-main.287",
    pages = "3905--3920"
}

@inproceedings{es-etal-2024-ragas,
    title = "{RAGA}s: Automated Evaluation of Retrieval Augmented Generation",
    author = "Es, Shahul  and
      James, Jithin  and
      Espinosa Anke, Luis  and
      Schockaert, Steven",
    editor = "Aletras, Nikolaos  and
      De Clercq, Orphee",
    booktitle = "Proceedings of the 18th Conference of the European Chapter of the Association for Computational Linguistics: System Demonstrations",
    month = mar,
    year = "2024",
    address = "St. Julians, Malta",
    publisher = "Association for Computational Linguistics",
    url = "https://aclanthology.org/2024.eacl-demo.16/",
    doi = "10.18653/v1/2024.eacl-demo.16",
    pages = "150--158"
}

@inproceedings{cheng2024adapting,
title={Adapting Large Language Models via Reading Comprehension},
author={Daixuan Cheng and Shaohan Huang and Furu Wei},
booktitle={The Twelfth International Conference on Learning Representations},
year={2024},
url={https://openreview.net/forum?id=y886UXPEZ0}
}

@inproceedings{tian2024finetuning,
title={Fine-Tuning Language Models for Factuality},
author={Katherine Tian and Eric Mitchell and Huaxiu Yao and Christopher D Manning and Chelsea Finn},
booktitle={The Twelfth International Conference on Learning Representations},
year={2024},
url={https://openreview.net/forum?id=WPZ2yPag4K}
}

@inproceedings{ovadia-etal-2024-fine,
    title = "Fine-Tuning or Retrieval? Comparing Knowledge Injection in {LLM}s",
    author = "Ovadia, Oded  and
      Brief, Menachem  and
      Mishaeli, Moshik  and
      Elisha, Oren",
    editor = "Al-Onaizan, Yaser  and
      Bansal, Mohit  and
      Chen, Yun-Nung",
    booktitle = "Proceedings of the 2024 Conference on Empirical Methods in Natural Language Processing",
    month = nov,
    year = "2024",
    address = "Miami, Florida, USA",
    publisher = "Association for Computational Linguistics",
    url = "https://aclanthology.org/2024.emnlp-main.15/",
    doi = "10.18653/v1/2024.emnlp-main.15",
    pages = "237--250"
}

@inproceedings{min-etal-2023-factscore,
    title = "{FA}ct{S}core: Fine-grained Atomic Evaluation of Factual Precision in Long Form Text Generation",
    author = "Min, Sewon  and
      Krishna, Kalpesh  and
      Lyu, Xinxi  and
      Lewis, Mike  and
      Yih, Wen-tau  and
      Koh, Pang  and
      Iyyer, Mohit  and
      Zettlemoyer, Luke  and
      Hajishirzi, Hannaneh",
    editor = "Bouamor, Houda  and
      Pino, Juan  and
      Bali, Kalika",
    booktitle = "Proceedings of the 2023 Conference on Empirical Methods in Natural Language Processing",
    month = dec,
    year = "2023",
    address = "Singapore",
    publisher = "Association for Computational Linguistics",
    url = "https://aclanthology.org/2023.emnlp-main.741/",
    doi = "10.18653/v1/2023.emnlp-main.741",
    pages = "12076--12100"
}

@article{wei2024long,
  title={Long-form factuality in large language models},
  author={Wei, Jerry and Yang, Chengrun and Song, Xinying and Lu, Yifeng and Hu, Nathan and Huang, Jie and Tran, Dustin and Peng, Daiyi and Liu, Ruibo and Huang, Da and others},
  journal={Advances in Neural Information Processing Systems},
  volume={37},
  pages={80756--80827},
  year={2024}
}

@article{ru2024ragchecker,
  title={Ragchecker: A fine-grained framework for diagnosing retrieval-augmented generation},
  author={Ru, Dongyu and Qiu, Lin and Hu, Xiangkun and Zhang, Tianhang and Shi, Peng and Chang, Shuaichen and Jiayang, Cheng and Wang, Cunxiang and Sun, Shichao and Li, Huanyu and others},
  journal={Advances in Neural Information Processing Systems},
  volume={37},
  pages={21999--22027},
  year={2024}
}

@inproceedings{pradeep2025great,
author = {Pradeep, Ronak and Thakur, Nandan and Upadhyay, Shivani and Campos, Daniel and Craswell, Nick and Soboroff, Ian and Dang, Hoa Trang and Lin, Jimmy},
title = {The Great Nugget Recall: Automating Fact Extraction and RAG Evaluation with Large Language Models},
year = {2025},
isbn = {9798400715921},
publisher = {Association for Computing Machinery},
address = {New York, NY, USA},
url = {https://doi.org/10.1145/3726302.3730090},
doi = {10.1145/3726302.3730090},
booktitle = {Proceedings of the 48th International ACM SIGIR Conference on Research and Development in Information Retrieval},
pages = {180–190},
numpages = {11},
location = {Padua, Italy},
series = {SIGIR '25}
}

@article{
kujanp2025efficient,
title={Efficient Knowledge Injection in {LLM}s via Self-Distillation},
author={Kalle Kujanp{\"a}{\"a} and Pekka Marttinen and Harri Valpola and Alexander Ilin},
journal={Transactions on Machine Learning Research},
issn={2835-8856},
year={2025},
url={https://openreview.net/forum?id=drYpdSnRJk},
note={}
}

@inproceedings{thorne-etal-2018-fever,
    title = "{FEVER}: a Large-scale Dataset for Fact Extraction and {VER}ification",
    author = "Thorne, James  and
      Vlachos, Andreas  and
      Christodoulopoulos, Christos  and
      Mittal, Arpit",
    editor = "Walker, Marilyn  and
      Ji, Heng  and
      Stent, Amanda",
    booktitle = "Proceedings of the 2018 Conference of the North {A}merican Chapter of the Association for Computational Linguistics: Human Language Technologies, Volume 1 (Long Papers)",
    month = jun,
    year = "2018",
    address = "New Orleans, Louisiana",
    publisher = "Association for Computational Linguistics",
    url = "https://aclanthology.org/N18-1074/",
    doi = "10.18653/v1/N18-1074",
    pages = "809--819"
}

@inproceedings{samarinas-etal-2025-beyond,
    title = "Beyond Factual Accuracy: Evaluating Coverage of Diverse Factual Information in Long-form Text Generation",
    author = "Samarinas, Chris  and
      Krubner, Alexander  and
      Salemi, Alireza  and
      Kim, Youngwoo  and
      Zamani, Hamed",
    editor = "Che, Wanxiang  and
      Nabende, Joyce  and
      Shutova, Ekaterina  and
      Pilehvar, Mohammad Taher",
    booktitle = "Findings of the Association for Computational Linguistics: ACL 2025",
    month = jul,
    year = "2025",
    address = "Vienna, Austria",
    publisher = "Association for Computational Linguistics",
    url = "https://aclanthology.org/2025.findings-acl.693/",
    doi = "10.18653/v1/2025.findings-acl.693",
    pages = "13468--13482",
    ISBN = "979-8-89176-256-5"
}

@inproceedings{hu-etal-2024-knowledge,
    title = "Knowledge-Centric Hallucination Detection",
    author = "Hu, Xiangkun  and
      Ru, Dongyu  and
      Qiu, Lin  and
      Guo, Qipeng  and
      Zhang, Tianhang  and
      Xu, Yang  and
      Luo, Yun  and
      Liu, Pengfei  and
      Zhang, Yue  and
      Zhang, Zheng",
    editor = "Al-Onaizan, Yaser  and
      Bansal, Mohit  and
      Chen, Yun-Nung",
    booktitle = "Proceedings of the 2024 Conference on Empirical Methods in Natural Language Processing",
    month = nov,
    year = "2024",
    address = "Miami, Florida, USA",
    publisher = "Association for Computational Linguistics",
    url = "https://aclanthology.org/2024.emnlp-main.395/",
    doi = "10.18653/v1/2024.emnlp-main.395",
    pages = "6953--6975"
}

@article{kappainterpret,
 ISSN = {0006341X, 15410420},
 URL = {http://www.jstor.org/stable/2529310},
 author = {J. Richard Landis and Gary G. Koch},
 journal = {Biometrics},
 number = {1},
 pages = {159--174},
 publisher = {International Biometric Society},
 title = {The Measurement of Observer Agreement for Categorical Data},
 urldate = {2026-06-06},
 volume = {33},
 year = {1977}
}

@inproceedings{metropolitansky-larson-2025-towards,
    title = "Towards Effective Extraction and Evaluation of Factual Claims",
    author = "Metropolitansky, Dasha  and
      Larson, Jonathan",
    editor = "Che, Wanxiang  and
      Nabende, Joyce  and
      Shutova, Ekaterina  and
      Pilehvar, Mohammad Taher",
    booktitle = "Proceedings of the 63rd Annual Meeting of the Association for Computational Linguistics (Volume 1: Long Papers)",
    month = jul,
    year = "2025",
    address = "Vienna, Austria",
    publisher = "Association for Computational Linguistics",
    url = "https://aclanthology.org/2025.acl-long.348/",
    doi = "10.18653/v1/2025.acl-long.348",
    pages = "6996--7045",
    ISBN = "979-8-89176-251-0"
}

@article{lin2025resisting,
  title={Resisting Contextual Interference in RAG via Parametric-Knowledge Reinforcement},
  author={Lin, Chenyu and Wen, Yilin and Su, Du and Tan, Hexiang and Sun, Fei and Chen, Muhan and Bao, Chenfu and Lyu, Zhonghou},
  journal={arXiv preprint arXiv:2506.05154},
  year={2025}
}

@inproceedings{kaster-etal-2021-global,
    title = "Global Explainability of {BERT}-Based Evaluation Metrics by Disentangling along Linguistic Factors",
    author = "Kaster, Marvin  and
      Zhao, Wei  and
      Eger, Steffen",
    editor = "Moens, Marie-Francine  and
      Huang, Xuanjing  and
      Specia, Lucia  and
      Yih, Scott Wen-tau",
    booktitle = "Proceedings of the 2021 Conference on Empirical Methods in Natural Language Processing",
    month = nov,
    year = "2021",
    address = "Online and Punta Cana, Dominican Republic",
    publisher = "Association for Computational Linguistics",
    url = "https://aclanthology.org/2021.emnlp-main.701/",
    doi = "10.18653/v1/2021.emnlp-main.701",
    pages = "8912--8925"
}

@article{olmo2025olmo,
  title={Olmo 3},
  author={Olmo, Team and Ettinger, Allyson and Bertsch, Amanda and Kuehl, Bailey and Graham, David and Heineman, David and Groeneveld, Dirk and Brahman, Faeze and Timbers, Finbarr and Ivison, Hamish and others},
  journal={arXiv preprint arXiv:2512.13961},
  year={2025}
}

@article{Yang2024Qwen25TR,
  title={Qwen2.5 Technical Report},
  author={Qwen An Yang and Baosong Yang and Beichen Zhang and Binyuan Hui and Bo Zheng and Bowen Yu and Chengyuan Li and Dayiheng Liu and Fei Huang and Guanting Dong and Haoran Wei and Huan Lin and Jian Yang and Jianhong Tu and Jianwei Zhang and Jianxin Yang and Jiaxin Yang and Jingren Zhou and Junyang Lin and Kai Dang and Keming Lu and Keqin Bao and Kexin Yang and Le Yu and Mei Li and Mingfeng Xue and Pei Zhang and Qin Zhu and Rui Men and Runji Lin and Tianhao Li and Tingyu Xia and Xingzhang Ren and Xuancheng Ren and Yang Fan and Yang Su and Yi-Chao Zhang and Yunyang Wan and Yuqi Liu and Zeyu Cui and Zhenru Zhang and Zihan Qiu and Shanghaoran Quan and Zekun Wang},
  journal={arXiv preprint arXiv:2412.15115},
  year={2024},
}


\appendix\label{sec:appendix}

\begin{figure*}[htpb]
\centering
\fbox{\includegraphics[width=\textwidth]{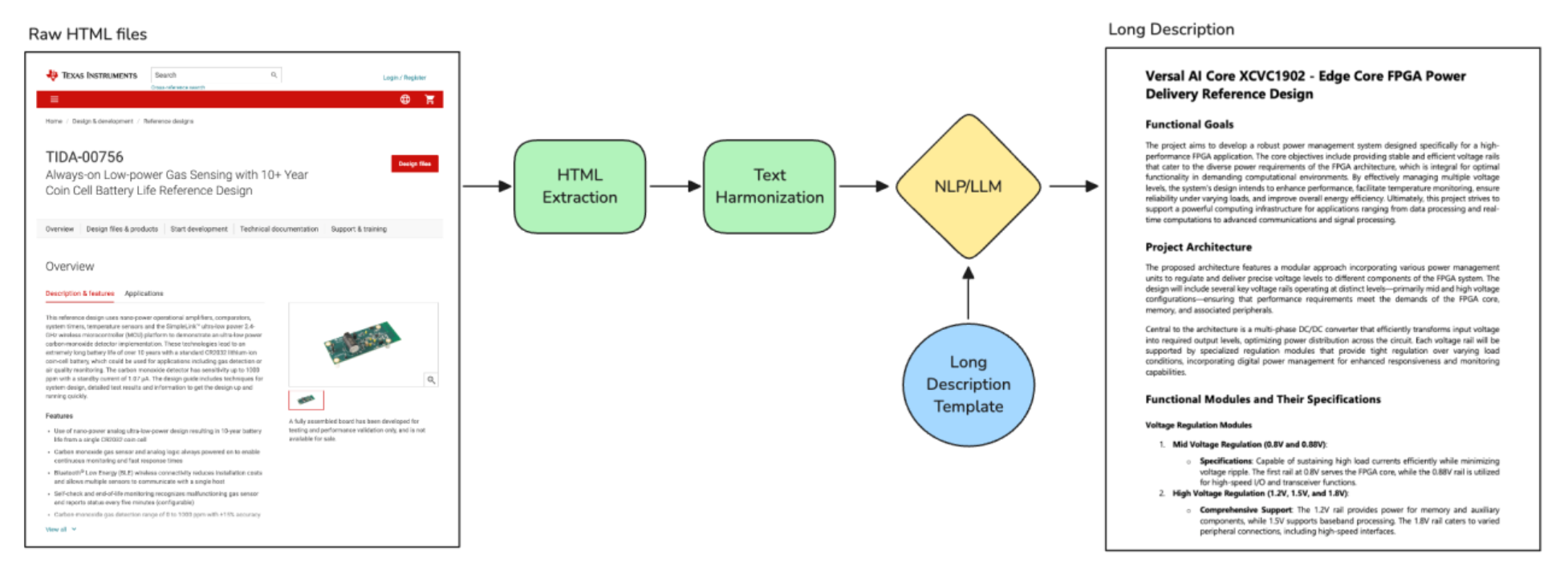}}
\caption{Our data creation pipeline: We first extract text from the HTML-files, then prompt GPT-4o to 1) harmonize the text, which means to take any manufacturer-specific text (e.g., component ids, project titles) and change it to fictional nouns, and 2) to use a specific template for generating a structured format, with mandatory sections (e.g., title, functional goals, project architecture).}
\label{fig:data_pipeline}
\end{figure*}


\begin{figure*}[htpb]
\centering
\fbox{\includegraphics[width=\textwidth]{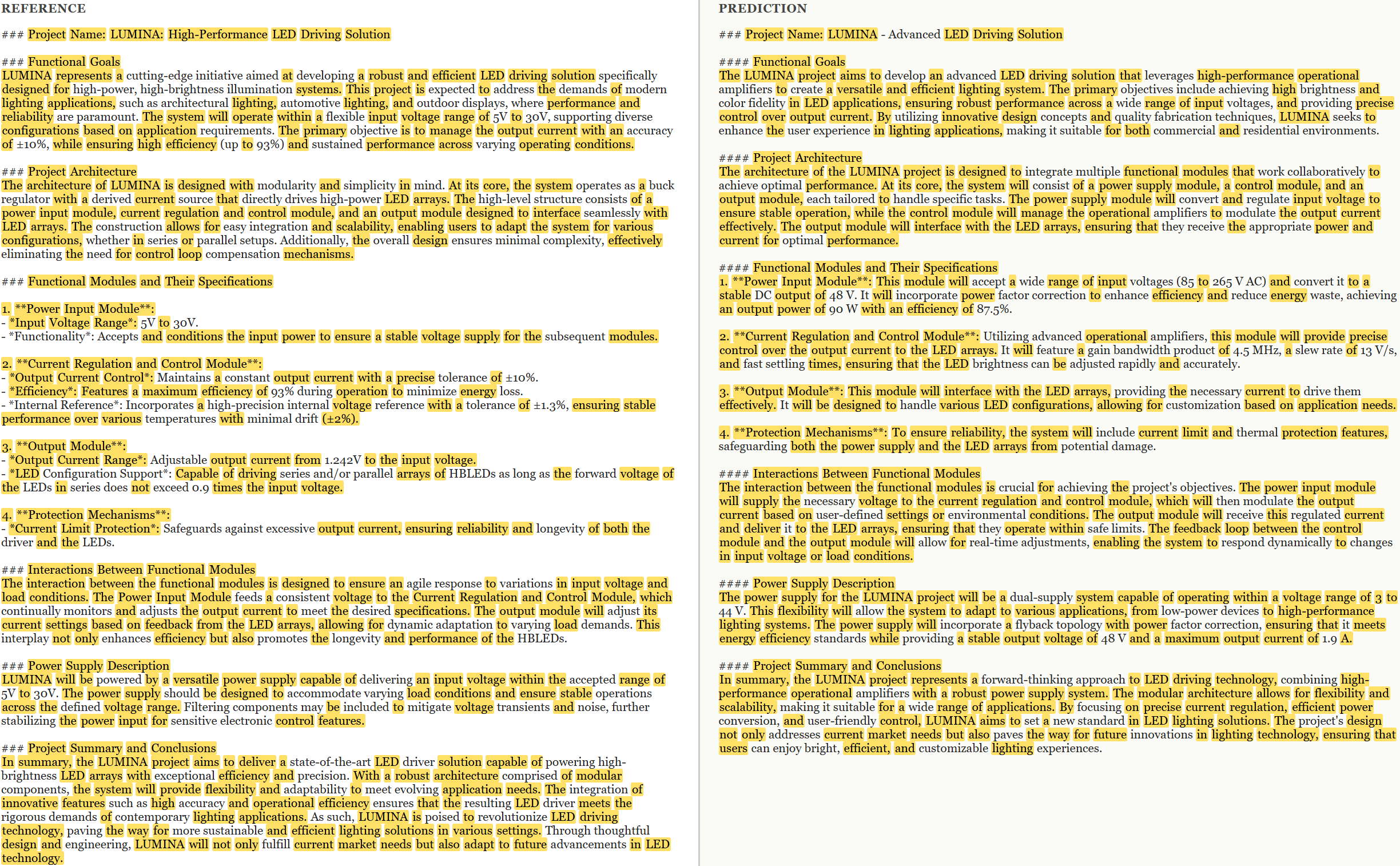}}
\caption{A model prediction, where Rouge-1 and claim-based Rec$_{ref}$ are disagreeing considerably: Rouge-1 states a high word overlap with the reference, but Rec$_{ref}$ shows very low reference-backed factuality. We see that almost every technical value in the predictions is not grounded in the reference (e.g. "input voltages (85 to 265 V AC)" "DC output of 48 V", "output power of 90 W", "gain bandwidth product of 4.5 MHz", "slew rate of 13 V/s"), but the predicted words and overall structure is very similar (reference: 454/640 tokens overlap ($70.9\%$); prediction: 477/644 tokens overlap ($74.1\%$)). Those hallucinations, well hidden under domain-specific prose, can by and large lead to false conclusions---for instance, interpreting an increase in Rouge-1 after training as a successful run.}
\label{fig:phen_rouge}
\end{figure*}

\begin{figure*}[t]
\centering
\begin{subfigure}{0.48\linewidth}
    \centering
    \includegraphics[width=\linewidth]{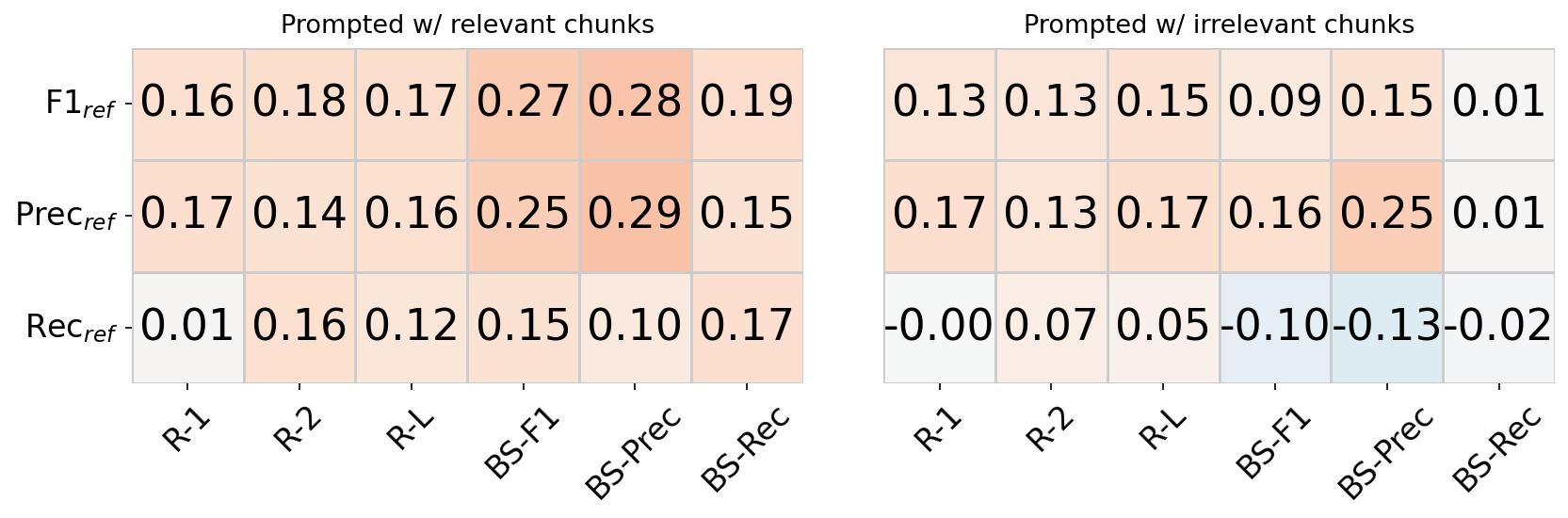}
    \caption{Olmo-3-7B}

\end{subfigure}

\vspace{\baselineskip}

\begin{subfigure}{0.48\linewidth}
    \centering
    \includegraphics[width=\linewidth]{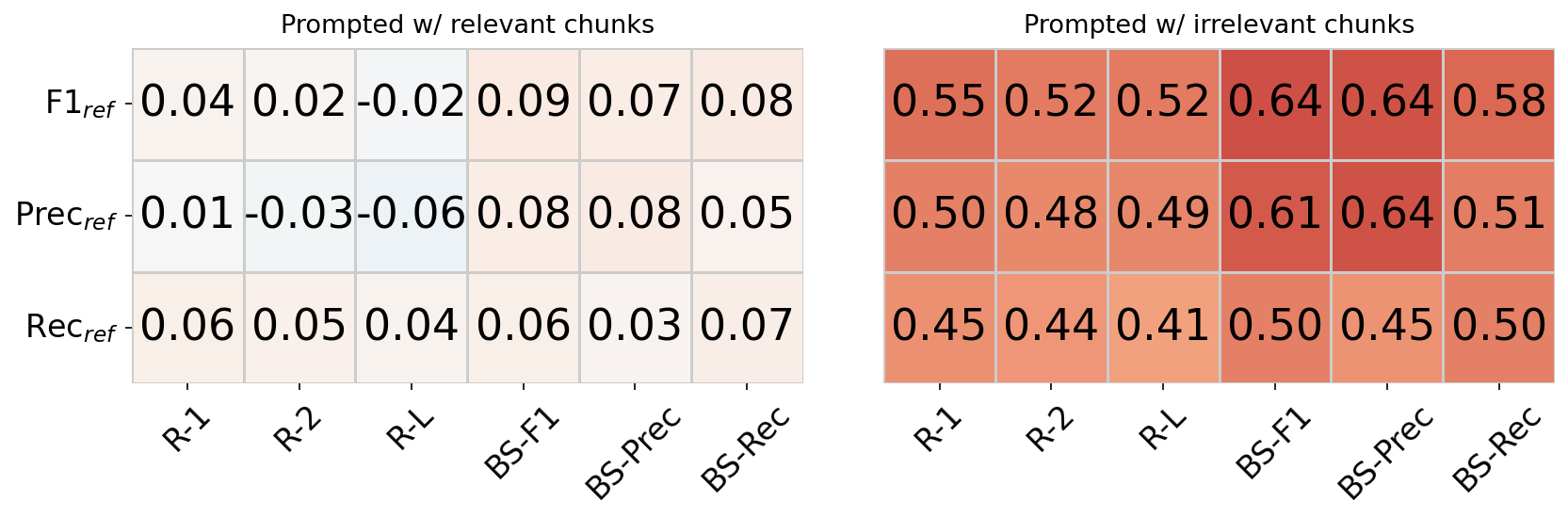}
    \caption{Olmo-3-7B + LoRA-base}

\end{subfigure}
\hfill
\begin{subfigure}{0.48\linewidth}
    \centering
    \includegraphics[width=\linewidth]{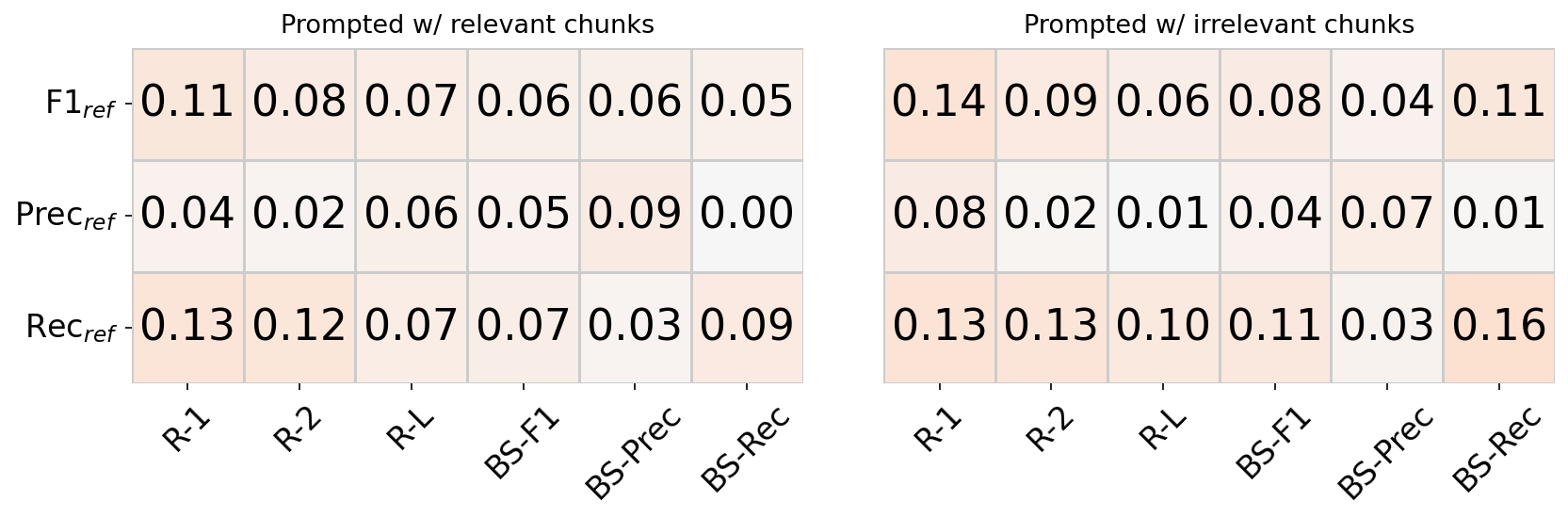}
    \caption{Olmo-3-7B + LoRA-irr}

\end{subfigure}

\caption{Spearman Rank Correlation of Rouge (R), BertScore (BS) and claim-based metrics.}
\label{fig:correlations-olmo}
\end{figure*}
\begin{figure*}[htpb]
\centering
\includegraphics[width=\textwidth]{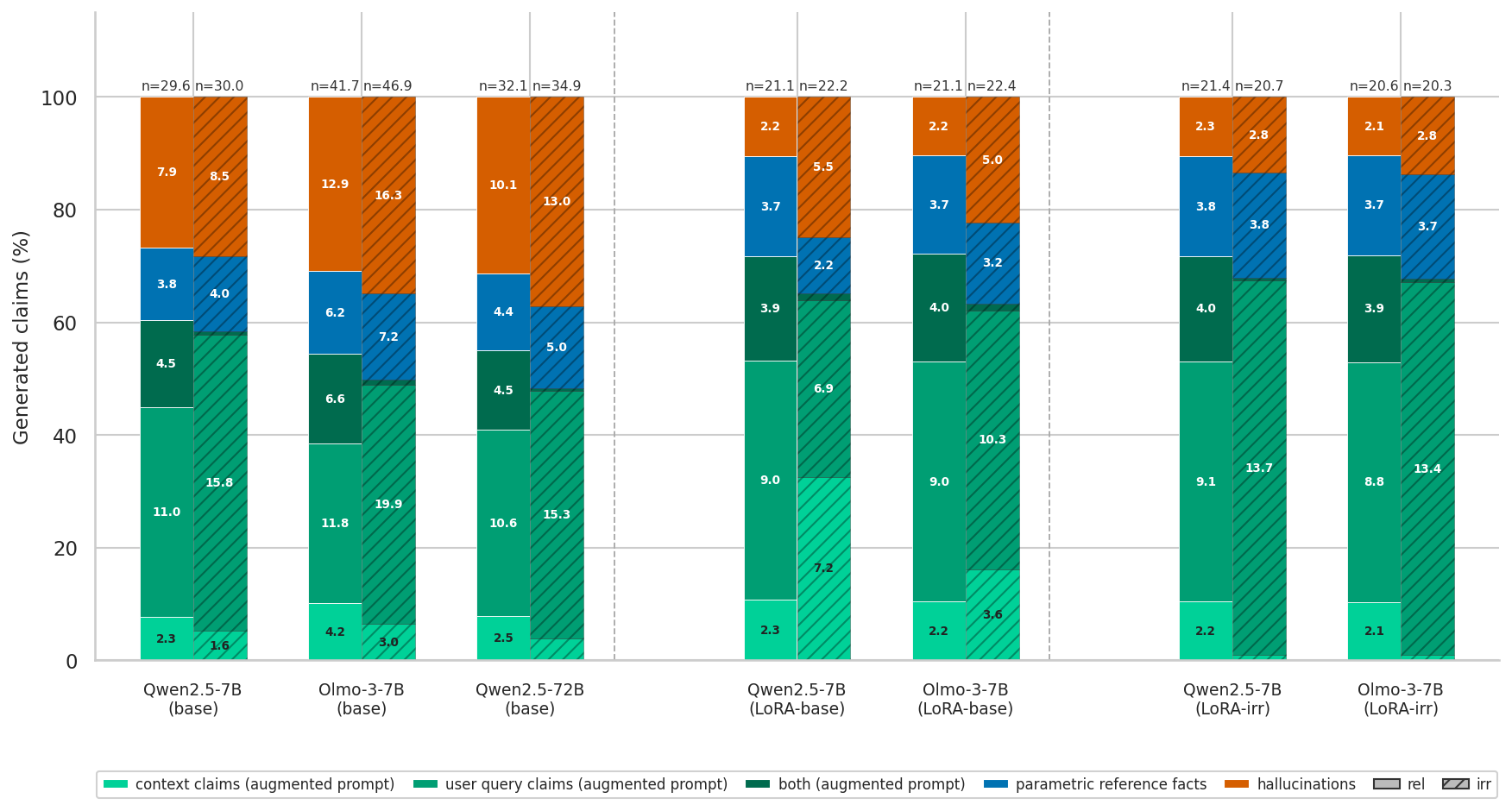}
\caption{Attribution of response claims to either the augmented prompt (context and user query), the reference response (parametric reference facts) or to no source at all (hallucinations). On top are the average claims per response (e.g. Qwen2.5-7B (base) generates on average approx. 30 claims). The labels annotating each source fraction report the mean claims per-example for each source (e.g. Qwen2.5-7B (base) generates on average approx. 4 parametric reference claims).}
\label{fig:source-distri}
\end{figure*}
\end{document}